%% file: 00-main.tex
\documentclass[10pt,journal,compsoc]{IEEEtran}

\usepackage{graphicx}
\usepackage{multirow}
\usepackage{hhline}
\usepackage{comment}
\usepackage{bigstrut}
\usepackage[dvipsnames]{xcolor}
\usepackage[sort&compress,square,comma,numbers]{natbib}
\usepackage[colorlinks=false, linkcolor=blue, citecolor=blue, bookmarks=true,pagebackref=false]{hyperref}
\usepackage{hyphenat}
\usepackage{subcaption}
\usepackage{float}
\usepackage{fancybox}
\usepackage[flushleft]{threeparttable}
\usepackage{array,booktabs}
\usepackage{amsmath}
\usepackage{outlines}
\usepackage[linesnumbered,ruled]{algorithm2e}
\usepackage[misc]{ifsym}
\usepackage{arydshln}
\usepackage{balance}
\usepackage{amsmath}
\usepackage{MnSymbol}
\usepackage{wasysym}
\usepackage{enumitem}
\usepackage{times}
\usepackage{fancyhdr}
\usepackage{courier}
\usepackage{booktabs}
\usepackage{pifont}
\usepackage{color,soul} 
\usepackage{tcolorbox}

\newcommand{\tastycolorbox}[1]{\begin{tcolorbox}[left=4pt,right=4pt,top=4pt,bottom=4pt,
colback=gray!5,colframe=gray!40!black,before skip=0pt,after skip=0pt]#1\end{tcolorbox}}

\usepackage{array}
\newcolumntype{R}[1]{>{\raggedleft\let\newline\\\arraybackslash\hspace{2pt}}m{#1}}

\makeatletter
\IEEEtriggercmd{\reset@font\normalfont\fontsize{7pt}{7pt}\selectfont}
\makeatother
\IEEEtriggeratref{1}

\newcommand{\ahmed}[1]{\textcolor{red}{{\it [Ahmed says: #1]}}}

\newcommand{\etal}[1]{${\textit{et al.}}$}

\newcolumntype{M}[1]{>{\centering\arraybackslash}m{#1}}

\newcommand\RQOne{For a given classifier, how much do the computed feature importance ranks by CA and CS methods differ across datasets?}
\newcommand\RQTwo{For a given dataset and classifier, how much do the computed feature importance ranks by the different CA methods differ?}
\newcommand\RQThree{On a given dataset, how much do the computed feature importance ranks by different CS methods differ?}

\ifCLASSINFOpdf
\else
\fi

\begin{document}
\bstctlcite{IEEEexample:BSTcontrol}
%
\title{The impact of feature importance methods on the interpretation of defect classifiers}

%
%
%
%

\author{Gopi~Krishnan~Rajbahadur,
Shaowei~Wang,
Gustavo~A.~Oliva,\\
Yasutaka~Kamei, and
Ahmed~E.~Hassan
\IEEEcompsocitemizethanks{
\IEEEcompsocthanksitem Gopi~Krishnan~Rajbahadur is with the Centre for Software Excellence, Huawei, Canada. 

Email:gopi.krishnan.rajbahadur1@huawei.com

\IEEEcompsocthanksitem Gustavo A. Oliva, and Ahmed~E.~Hassan are with Software Analysis and Intelligence Lab (SAIL), School of Computing, Queen's University, Canada.\protect\\
E-mail: $\{$gustavo, ahmed$\}$@cs.queensu.ca

\IEEEcompsocthanksitem Shaowei~Wang is with the department of computer science, University of Manitoba, Canada.

Email: shaowei@cs.umanitoba.ca

\IEEEcompsocthanksitem Yasutaka~Kamei is with Principles of Software Languages (POSL) Lab, Graduate School and Faulty of Information Science and Electrical Engineering, Kyushu University, Japan.

Email: kamei@ait.kyushu-u.ac.jp

\IEEEcompsocthanksitem Shaowei Wang is the corresponding author

}

}

\IEEEtitleabstractindextext{%
\begin{abstract}
\input{01-abstract}

\end{abstract}

}

\maketitle

\IEEEdisplaynontitleabstractindextext

\IEEEpeerreviewmaketitle

\vspace{-2cm}

\input{02-introduction.tex}
\input{03-motivation-related_work.tex}

\input{04-casestudy-setup.tex}
\input{05-casestudy-results.tex}

\input{06-discussion.tex}
\input{07-implications.tex}
\input{08-threats.tex}
\input{09-conclusion.tex}

\ifCLASSOPTIONcaptionsoff
  \newpage
\fi



%

\bibliographystyle{IEEEtranN}

\begin{footnotesize}
    \bibliography{main}
\end{footnotesize}

\input{bios}
	
\begin{appendices}
	\input{appendix.tex}		
\end{appendices}

\end{document}

%% file: 01-abstract.tex
Classifier specific (CS) and classifier agnostic (CA) feature importance methods are widely used (often interchangeably) by prior studies to derive feature importance ranks from a defect classifier. However, different feature importance methods are likely to compute different feature importance ranks even for the same dataset and classifier. Hence such interchangeable use of feature importance methods can lead to conclusion instabilities unless there is a strong agreement among different methods. Therefore, in this paper, we evaluate the agreement between the feature importance ranks associated with the studied classifiers through a case study of 18 software projects and six commonly used classifiers. We find that: 1) The computed feature importance ranks by CA and CS methods do not always strongly agree with each other. 2) The computed feature importance ranks by the studied CA methods exhibit a strong agreement including the features reported at top-1 and top-3 ranks for a given dataset and classifier, while even the commonly used CS methods yield vastly different feature importance ranks. Such findings raise concerns about the stability of conclusions across replicated studies.  
We further observe that the commonly used defect datasets are rife with feature interactions and these feature interactions impact the computed feature importance ranks of the CS methods (not the CA methods). 
We demonstrate that removing these feature interactions, even with simple methods like CFS improves agreement between the computed feature importance ranks of CA and CS methods. 
In light of our findings, we provide guidelines for stakeholders and practitioners when performing model interpretation and directions for future research, e.g., future research is needed to investigate the impact of advanced feature interaction removal methods on computed feature importance ranks of different CS methods.

%% file: 02-introduction.tex
\section{Introduction}
\label{sec:introduction}

Defect classifiers are widely used by many large software corporations~\cite{lewis2013does,zimmermann2009cross,caglayan2015merits,shihab2012industrial} and researchers~\cite{zhang2016use,shihab2011high,chen2018applications}. Defect classifiers are commonly interpreted to uncover insights to improve software quality. Such insights help practitioners formulate strategies for effective testing, defect avoidance, and quality assurance~\cite{jiarpakdee2019impact,theisen2015approximating}. Therefore it is pivotal that these generated insights are reliable.

When interpreting classifiers, prior studies typically employ a feature importance method to compute a ranking of feature importances (a.k.a., feature importance ranks)~\cite{jiarpakdee2019impact,herzig2016impact,guo2004robust,jahanshahi2019does,mori2019balancing}. These feature importance ranks reflect the order in which the studied features contribute to the predictive capability of the studied classifier~\cite{hooker2019benchmark}. These feature importance methods can be divided in two categories: classifier-specific (CS) and classifier-agnostic (CA) methods. A classifier-specific (CS) method makes use of a given classifier's internals to measure the degree to which each feature contributes to a classifier's predictions~\cite{molnar2018interpretable}. 
We note, however, that a CS method is not always readily available for a given classifier. For example, complex classifiers like SVMs and deep neural networks do not have a widely accepted CS method(s)~\cite{chakraborty2017interpretability}.

For cases such as those, or when a universal way of comparing the feature importance ranks of different classifiers is required~\cite{rajbahadur2017impact,tantithamthavorn2018impact}, classifier-agnostic (CA) methods are typically used. Such CA methods measure the contribution of each feature towards a classifier’s predictions. For instance, some CA methods measure the contribution of each feature by effecting changes to that particular feature in the dataset and observing its impact on the outcome. The primary advantage of CA methods is that they can be used for any classifier (i.e., from interpretable to black box classifiers).



Despite computing feature importance ranks using different ways, CS and CA methods are indiscriminately and interchangeably used in software engineering studies (Table \ref{tab:motivation}). For instance, to compute feature importance ranks for a random forest classifier,~\citet{treude2019predicting} and \citet{yu2018conpredictor} use CS methods: the Gini importance and the Breiman’s importance methods respectively. On the other hand, \citet{mori2019balancing} and \citet{herzig2014using} use CA methods: Partial Dependence Plot (PDP) and filterVarImp respectively. Since these methods compute feature importances differently (Section~\ref{sec:fi-methods}), different CS or CA methods are likely to compute different feature importance ranks for the same classifier. Yet, we observe that the rationale for choosing a given feature importance method is rarely motivated by prior studies (Section \ref{sec:motivation}).

The interchangeable use of feature importance methods is acceptable only if the feature importance ranks computed by these methods do not differ from each other. Therefore, in order to determine the extent to which the importance ranks computed by different importance methods agree with each other, we conduct a case study on 18 popularly used software defect datasets using classifiers from six different families. We compute the feature importance ranks using six CS and two CA methods on these datasets and classifiers. The list of CS methods is summarized in Table~\ref{tab:approach_classifiers}. The two CA methods are: permutation importance (Permutation) and SHapley Additive ExPlanations (SHAP). Finally, we compute Kendall's Tau, Kendall's W, and Top-k ($k \in \{1,3\}$) overlap to quantify the agreement between the computed feature importance ranks by the different studied feature importance methods for a given classifier and dataset. While Kendall's measures compute differences across the different feature importance ranks, the Top-K overlap measure focuses on the top-k items of these rankings (more details in Section \ref{sec:metrics}). 
We highlight our findings below:

\begin{itemize}
 \item The computed feature importance ranks by CA and CS methods do not always strongly agree with each other. For two of the five studied classifiers, even the most important feature varies across CA and CS methods.
 \item The computed feature importance ranks by the studied CA methods exhibit a strong agreement including the features reported at top-1 and top-3 ranks for a given dataset and classifier.
 \item On a given dataset, even the commonly used CS methods yield vastly different feature importance ranks, including the top-3 and the top-1 most important feature(s).
\end{itemize}

We then investigate why the agreement between the different CS methods and CA and CS methods remains weak, while the agreement between the computed feature importance ranks by CA methods is strong. Through a simulation study we find that, as hinted by prior studies~\cite{de2007interpretation, freeman1985analysis, fisher2018all,devlin2019disentangled,lundberg2018consistent}, feature interactions present in the studied datasets impact the computed feature importance ranks of CS methods. We then investigate if removal of feature interaction in a given dataset (through a simple method like Correlation-based Feature Selection (CFS)~\cite{hall1999correlation,hall2003benchmarking}) improves the agreement between the computed feature importance ranks of studied feature importance methods. We find that removal of feature interaction, significantly improves the agreement between the computed feature importance ranks of CA and CS methods. However, the improvement in agreement between the computed feature importance ranks of the studied CS methods remains marginal. In light of these findings, we suggest that future research on defect classification should:


\begin{enumerate}[wide = 0pt, itemsep = 3pt]
    \item Identify (e.g., using methods like Friedman’s H-statistic (more details in Section~\ref{sec:disc2})) and remove feature interactions present in the dataset (e.g., using simple methods like CFS) before interpreting the classifier, as they hinder the classifier interpretation.

    \item One should always specify the used feature importance method to increase the reproducibility of their study and the generalizability of its insights.


\end{enumerate}


\noindent \textbf{Paper Organization.} Section~\ref{sec:motivation_related} presents the motivation of our study and the related work. Section~\ref{sec:casestudy} explains how we conducted our case study. In Section~\ref{sec:rqs}, we present the results of our case study which examines the extent to which the feature importance ranks computed by different feature importance methods vary. In Section~\ref{sec:disc}, we investigate the impact of feature interactions on the computed feature importance ranks by studied interpretation methods. Section~\ref{sec:implications} presents the implications of our results and avenues for future research. Section~\ref{sec:threat} lists the threats to the validity of our study. Finally, Section~\ref{sec:conclusion} concludes our study.


%% file: 03-motivation-related_work.tex
\section{Motivation and Related Work}\label{sec:motivation_related}

In this section, we motivate our study based on how prior studies employed feature importance methods (Section \ref{sec:motivation}). Next, we situate our study relative to prior related work (Section \ref{sec:related}).

\subsection{Motivation}
\label{sec:motivation}

\begin{table*}[htbp]
  \centering
  \caption{Different feature importance methods used for interpreting various classifiers in the software engineering literature}
  \begin{threeparttable}
    \begin{tabular}{p{2.5cm}|p{4cm}|p{3cm}|p{2.5cm}|p{2.7cm}}
    \hline
    \textbf{Classifier Family} & \textbf{Papers using CS} & \textbf{Used CS methods} & \textbf{Papers using CA} & \textbf{Used CA methods} \\
    \hline
    \textbf{Statistical Techniques} & \cite{subramanyam2003empirical}\cite{herbsleb2003empirical}\textsuperscript{\ding{51}}\cite{zimmermann2008predicting, angelis2001building,mori2019balancing,nagappan2005use,morales2015code,ghaleb2019empirical,kononenko2015investigating}& Regression coefficients, ANOVA & \cite{calefato2019empirical,herzig2014using,herzig2016impact,premraj2011network} & Boruta\textsuperscript{\ding{83}}, filterVarImp\textsuperscript{\ding{61}}\\
    \hline
    \textbf{Rule-Based Techniques} & \cite{gay2010automatically}\textsuperscript{\ding{51}},\cite{othmane2017time} & Interpreting rules, varImp\textsuperscript{\ding{39}} & \cite{calefato2019empirical} & Boruta\textsuperscript{\ding{83}} \\
    \hline
    \textbf{Neural Networks} & \cite{santos2013estimating}\textsuperscript{\ding{54}}~\cite{ma2018mode}\textsuperscript{\ding{51}} & MODE\textsuperscript{\cite{ma2018mode}} & \cite{calefato2019empirical} & Boruta\textsuperscript{\ding{83}} \\
    \hline
    \textbf{Decision Trees} & \cite{el2001modelling}\cite{knab2006predicting}\textsuperscript{\ding{51}}\cite{malgonde2019ensemble} & Decision branches, Gini importance & \cite{calefato2019empirical,herzig2014using,herzig2016impact,premraj2011network} & Boruta\textsuperscript{\ding{83}}, filterVarImp\textsuperscript{\ding{61}} \\
    \hline
    \textbf{Ensemble methods- Bagging } & \cite{treude2019predicting,yu2018conpredictor,haran2007techniques}~\cite{guo2004robust}\textsuperscript{\ding{54}}~\cite{gousios2014exploratory,Niedermayr2019}~\cite{martens2019towards}\textsuperscript{\ding{54}}~\cite{fan2018early, bao2019large,jahanshahi2019does}~\cite{dey2018software}\textsuperscript{\ding{54}} & Permutation importance, Gini importance & \cite{mori2019balancing,calefato2019empirical,herzig2014using,herzig2016impact,premraj2011network,dehghan2017predicting,blincoe2019high} & Boruta\textsuperscript{\ding{83}}, filterVarImp\textsuperscript{\ding{61}}, PDP, Marks method\textsuperscript{\cite{marks2011studying}}, BestFirst\textsuperscript{\ding{72}} \\
    \hline
    \textbf{Ensemble methods- Boosting} &  -  &  -  & \cite{calefato2019empirical,herzig2014using,herzig2016impact} & Boruta\textsuperscript{\ding{83}}, filterVarImp\textsuperscript{\ding{61}} \\
    \hline
    \end{tabular}%
    \begin{tablenotes}
    \scriptsize
    \item \ding{54} - The used method for computing the feature importance ranks is not mentioned
    \item \ding{51} - Papers in which the rationale for choosing a given feature importance method is specified
    \item \ding{83} - \url{https://cran.r-project.org/web/packages/Boruta/index.html}
    \item \ding{61} - \url{https://www.rdocumentation.org/packages/caret/versions/6.0-84/topics/filterVarImp}
    \item \ding{72} - \url{https://www.rdocumentation.org/packages/FSelector/versions/0.31/topics/best.first.search}
    \item \ding{39} - \url{https://www.rdocumentation.org/packages/caret/versions/6.0-84/topics/varImp}
    
\end{tablenotes}
\end{threeparttable}
  \label{tab:motivation}%
\end{table*}%

We conduct a literature survey of the used feature importance methods in prior studies. To survey the literature, we searched Google Scholar with the terms ``software engineering", ``variable importance", ``feature importance'' and the name of each classifier that is studied in our paper (Section~\ref{sec:approach}). We searched the Google Scholar multiple times, once for each studied classifier. We eliminated all the papers that were from before the year 2000 to restrict the scope of our survey to recent studies. We read each paper from the search results in order to check if they employed any feature importance method(s) to generate insights. We consider all the studies presented in the google scholar and do not filter based on venues. However, we do not include the papers in which one of the authors of this current study was involved to avoid potential confirmation bias. A summary of our literature survey is shown in Table~\ref{tab:motivation}.

We observe that studies rarely specify the reason for choosing their feature importance method -- only four out of the 29 surveyed studies provide a rationale for choosing their used feature importance method. 

We note that both CA and CS methods are widely used. For instance, from Table~\ref{tab:motivation}, we see that both Gini importance and filterVarImp have been used to interpret a random forest classifier. However, given that (i) feature importance methods are typically used to generate insights and (ii) different methods compute the feature importance using different approaches, such interchangeable usage of methods on a given classifier in prior studies is troublesome~\cite{menzies2012special}. 


For instance, \citet{zimmermann2008predicting} used an F-Test on the coefficients of a logistic regression classifier (a CS method) to show that there exists a strong empirical relationship between Social Network Analysis (SNA) metrics and the defect proneness of a file. Later, \citet{premraj2011network} used filterVarImp (a CA method) and logistic regression classifier to show that empirical relationship between SNA metrics and the defective files are negligible. Given that CS and CA methods can produce different feature importance ranks, it is unclear whether the aforementioned conflicting result is due to absence of an empirical relationship in the data or simply due to the differing feature importance methods and as such leads to conclusion instability.

More generally, the interchangeable use of feature importance methods (i.e., CS and CA methods), when replicating a study, is acceptable only if the computed ranks by different methods are not vastly different on a given dataset. Otherwise, it raises concerns about the stability of conclusions across the replicated studies. Hence, we investigate the following research question: 

\smallskip \tastycolorbox{(RQ1) \RQOne} \smallskip

Similar concerns exist regarding the interchangeable use of different CA methods, even for the same classifier. From Table~\ref{tab:motivation}, we observe that, within each classifier family, different studies use different CA methods. The rationale for choosing a given CA method (for instance, filterVarImp) over another (for instance, PDP) is rarely provided. For instance, none of the studies using a CA method in Table~\ref{tab:motivation} provide reasons for choosing one CA method over another. Yet, the extent to which these CA agree with each other is unclear. Such a concern becomes particularly relevant with the recent rise of complex classifiers for defect prediction~\cite{dam2018explainable, chen2018applications,hihn2015data}, as these classifiers do not have a universally agreed-upon or popular CS method. Hence, we investigate the following research question:

\smallskip \tastycolorbox{(RQ2) \RQTwo} \smallskip

A number of general prior studies already note that feature importance ranks differ vastly between CS methods~\cite{strobl2007bias}. However, such a comparison among CS methods pertaining to different classifier (i.e., CS method associated with a decision tree classifier and a random forest classifier) has not been studied, in the context of defect prediction and software engineering. Such a study is extremely important to understand the limits of reproducibility and generalizability of prior studies. 
For instance, \citet{jahanshahi2019does} replicate the study of \citet{mcintosh2017fix} using random forest (and the CS methods of random forest classifier) as opposed to the non-linear logistic regression classifier and its associated CS method. \citet{jahanshahi2019does} observe that their feature importance ranks differ from those of the original study. In particular, different CS methods are likely to compute feature importances differently and the difference in insight could be attributed to the used CS method rather than the underlying phenomena (e.g., just-in-time defect prediction) that is being studied. Therefore, we study the following research question along with the previous ones:

\smallskip \tastycolorbox{(RQ3) \RQThree}

\subsection{Related Work}
\label{sec:related}
As summarized in Section~\ref{sec:motivation}, both CA and CS methods have been widely used by the software engineering researchers to compute feature importance ranks. In the following, we describe related work regarding (i) usage of feature importance methods in software engineering (ii) the problems associated with widely used feature importance methods and (iii) sensitivity of feature importance methods:

\smallskip \noindent \textbf{Usage of Feature Importance Methods in Software Engineering.} Both CA and CS methods have been widely used by software engineering researchers to compute feature importance ranks. For instance, \citet{mcintosh2016empirical} and \citet{morales2015code} construct regression models and use ANOVA (a CS method) to understand which aspects of code review impact software quality. In turn, \citet{fan2018early} use the CS methods that are associated with the random forest classifier to identify the features that distinguish between merged and abandoned code changes. Similarly, various CS methods that are associated with the random forest classifier have been used to identify features that are important for determining who will leave a company~\citep{bao2017will}, who will become a long time contributor to an open source project~\cite{bao2019large}, code metrics that signal defective code~\citep{guo2004robust}, popularity of a mobile app~\citep{tian2015characteristics}, likelihood of an issue being listed in software release notes~\citep{abebe2016empirical}. Furthermore, CS methods that are associated with logistic regression and decision trees have also been used to generate insights on similar themes~\citep{briand1998predicting, cataldo2009software, calefato2019empirical, gay2010automatically, bird2011don, bird2009does}. Correspondingly, previous studies also use CA methods to interpret classifiers. For example,  \citet{tantithamthavorn2018impact} and \citet{rajbahadur2017impact} use the permutation CA method to study the impact of data pre-processing on a classifier’s feature importance ranks. Furthermore, \citet{dey2018software} use partial dependence plots (PDP) to identify why certain metrics are not important for predicting the change popularity of an npm package, whereas \citet{mori2019balancing} use PDP to compute the feature importance of random forest classifiers. More recently,~\citet{jiarpakdee2020empirical} demonstrated how instance-level CA methods like LIME-HPO (Locally Interpretable Model-Agnostic Explanations with Hyperparameter Optimization) and Breakdown~\cite{Gosiewska2019ibreakdown} can be used to identify features that are important in determining whether a given module will become defective. They further show that these instance-level feature importances mostly agree with the traditional feature importance ranks.
 
\smallskip \noindent \textbf{Problems associated with widely used feature importance methods.} 
Prior studies investigated the potential problems and concerns regarding the widely used feature importance methods. \citet{strobl2007bias} find that CS methods associated with the widely used random forest classifier are biased and that different CS methods might yield different feature importance ranks when features are correlated. To deal with such correlations, \citet{strobl2008conditional} propose a conditional permutation importance method. Similarly, \citet{candes2018panning} propose ways to mitigate to potential false discoveries caused by the feature importance method of generalized linear models. \citet{lundberg2018consistent} and \citet{Gosiewska2019ibreakdown} identify that several popular CA methods produce imprecise feature importance ranks.

\smallskip \noindent \textbf{Sensitivity of feature importance methods.} While the aforementioned studies focus on finding potential problems with existing methods, very few studies compare the existing and widely used feature importance methods. For instance, \citet{gromping2009variable} compares the feature importance methods of linear regression and logistic regression and identifies both similarities and differences, whereas, \citet{auret2011empirical} compare several tree-based feature importance methods and find that CS method associated with conditional bagged inference trees to be robust. 

\citet{jiarpakdee2019impact} analyze the impact of correlated features on the feature importance ranks of a defect classifier. They find that including correlated features when building a defect classifier results in generation of inconsistent feature importance ranks. In order to avoid that, \citet{jiarpakdee2019impact} recommend practitioners to remove all the correlated features before building a defect classifier. Through a different study, \citet{jiarpakdee2018autospearman} propose an automated method called AutoSpearman that helps practitioners to automatically remove correlated and redundant features from a dataset. They demonstrate that the AutoSpearman method helps one to avoid the harmful impact of these correlated metrics on the computed feature importance ranks of a defect classifier. Similarly, ~\citet{tantithamthavorn2015impact} find that noise introduced in the defect datasets due to the mislabelling of defective modules influences the computed feature importance ranks. They show that, among the features reported in the top-3 ranks, the noise introduced by mislabelling does not impact the feature reported at rank 1. However, it influences the features reported at rank 2 and rank 3 across several defect classifiers that they study. 

Similarly, several prior studies from~\citet{tantithamthavorn2016empirical,tantithamthavorn2018impact,tantithamthavorn2018impact1} investigate how various experimental design choices impact the computed feature importance ranks of a classifier. For instance, ~\citet{tantithamthavorn2018impact} investigated whether class rebalancing methods impact the computed feature importance ranks of a classifier. They observe that using class rebalancing methods can introduce \textit{concept drift} in a defect dataset. This concept drift, in turn, impacts the computed feature importance ranks of a defect classifier constructed on the rebalanced defect dataset. ~\citet{tantithamthavorn2018impact1}also investigate the impact of hyperparameter optimization on the computed feature importance ranks of a classifier. They find that the features reported at the top-3 features importance ranks differ significantly between the hyperparameter tuned classifiers and untuned classifiers. \citet{rajbahadur2017impact}  investigate if the feature importance ranks computed for the regression-based random forest classifiers vary when computed by a CS method and Permutation CA method.


However, to the best of our knowledge, our study is the first work to empirically measure the agreement between the feature importance ranks computed by  CA and CS methods across 18 datasets, six classifiers, and 8 feature importance methods, especially in the context of defect prediction. Our study enables the software engineering community to assess the impact of using various feature importance methods interchangeably. As it raises concerns about the stability of conclusions across replicated studies -- as such conclusion instabilities might be due primarily to changes in the used feature importance methods instead of characteristics that are inherent in the domain.

%% file: 04-casestudy-setup.tex
\section{Case study setup}
\label{sec:casestudy}

In this section, we describe the studied datasets (Section~\ref{sec:data}), classifiers (Section~\ref{sec:classifiers}), and feature importance methods (Section~\ref{sec:fi-methods}). We then describe our case study approach (Section \ref{sec:approach}), as well as the evaluation metrics that we employ (Section \ref{sec:metrics}).

\input{04a-datasets}
\input{04b-classifiers.tex}
\input{04c-featureimportances.tex}
\input{04d-approach.tex}
\input{04e-metrics.tex}

%% file: 04a-datasets.tex
\subsection{Studied Datasets}
\label{sec:data}


We use the software project datasets from the PROMISE repository~\citep{Sayyad-Shirabad+Menzies:2005}. The data set contains the defect data of 101 software projects that are diverse in nature. Use of such varied software projects in our study helps us successfully mitigate the researcher bias identified by Shepperd et al. \citep{shepperd2014researcher,tantithamthavorn2016comments}. In addition, similar to the prior studies by \citet{rajbahadur2017impact} and \citet{tantithamthavorn2018impact}, we further filter the datasets to study based on two criteria. We remove the datasets with EPV less than 10 and the datasets with defective ratio less than 50. After filtering the 101 datasets from PROMISE with the aforementioned criteria, we end up with 18 datasets for our study: Poi-3.0, Camel-1.2, Xalan-2.5, Xalan-2.6, Eclipse34\_debug, Eclipse34\_swt, Pde, PC5, Mylyn, Eclipse-2.0, JM1, Eclipse-2.1, Prop-5, Prop-4, Prop-3, Eclipse-3.0, Prop-1, Prop-2. Table~\ref{tab:data} in Appendix \ref{appendix:studied-datasets} shows various basic characteristics about each of the studied datasets.

%% file: 04b-classifiers.tex
\subsection{Studied Classifiers}
\label{sec:classifiers}

We construct classifiers to evaluate our outlined research questions from Section~\ref{sec:motivation}. We choose the classifiers based on two criteria. First, the classifiers should be representative of the eight commonly used machine learning families in Software Engineering literature as given by \citet{ghotra2015revisiting}. The goal behind this criterion is to foster the generalizability and applicability of our results. Second, the chosen classifiers should have a CS method. We only choose classifiers with a CS method, so that we can compare and evaluate the computed feature importance ranks by the CA methods against the CS feature importance ranks for a given classifier (RQ1). In addition, such a choice enables us to compare the agreement between the computed feature importance ranks between different classifiers (RQ3). After the application of these criteria, we eliminate three machine learning families (clustering based classifiers, support vector machines, and nearest neighbour), as the classifiers from those families do not have a CS method. Furthermore, we split the ensemble methods family given by \citet{ghotra2015revisiting} into two categories to include classifiers belonging to both bagging and boosting families. The classifiers we finally choose are: Regularized Logistic Regression (glmnet), C5.0 Rule-Based Tree (C5.0Rules), Neural Networks (with model averaging) (avNNet), Recursive partitioning and Regression Trees (rpart), Random Forest (rf), Extreme Gradient Boosting Trees (xgbTree). Table~\ref{tab:approach_classifiers} in Appendix \ref{appendix:classifiers} shows the studied classifiers, their hyperparameters and machine learning families to which they belong.

We choose one representative classifier from each of the machine learning family from the \textbf{caret}\footnote{https://cran.r-project.org/web/packages/caret/index.html} package in R. Table~\ref{tab:approach_classifiers} also shows the caret function that was used to build the classifiers. The selected classifiers have a CS method, that is given by the \textbf{varImp()} function in the caret package.

Inherently interpretable classifiers (e.g., fast-and-frugal trees, naive-bayes classifiers and simple decision trees) do not benefit as much from feature importance methods. Hence, such classifiers are out of the scope of this study. Nevertheless, we strongly suggest that the future studies should also explore the reliability of the insights that is derived from simple interpretable classifiers.

%% file: 04c-featureimportances.tex
\subsection{Studied Feature Importance Methods}
\label{sec:fi-methods}

\subsubsection{Classifier Specific Feature Importance (CS) methods}

The CS methods typically make use of a given classifier's internals to compute the feature importance scores. These methods are widely used in software engineering to compute feature importance ranks as evidenced from Table~\ref{tab:motivation}.

We use six CS methods that are associated with the six classifiers that we study, namely: Logistic Regression FI (LRFI), C5.0 Rule-Based Tree FI (CRFI), Neural Networks (with model averaging) FI (NNFI), Recursive Partitioning and Regression Trees FI (RFI), Random Forest FI (RFFI), and  Extreme Gradient Boosting Trees FI (XGFI). Table~\ref{tab:CS} in Appendix \ref{appendix:cs-methods} provides a brief explanation about the inner working of these CS methods on a given classifier. For a more detailed explanation we refer the readers to~\citet{kuhn2012variable}.

\subsubsection{Classifier Agnostic Feature Importance (CA) methods}
\label{sec:CA}

CA methods compute the importance of a feature by treating the classifier as a “black-box”, i.e., without using any classifier-specific details. In this study, we use the permutation feature importance (Permutation) and SHapley Additive exPlanation (SHAP) CA methods. We use these two CA methods instead of others for the following reasons. First, Permutation is one of the oldest and most popularly used CA methods in both machine learning and software engineering communities~\cite{rajbahadur2017impact,tantithamthavorn2018impact,avati2018improving,janitza2013auc,jiarpakdee2019towards,husain2015understanding,jiarpakdee2020impact}. It was first introduced by~\citet{breiman2001random} as way of measuring the feature importance ranks of a random forest classifier and later was adopted as a CA method. Second, we consider SHAP, as it one of the more recent global feature importance method that is theoretically guaranteed to produce optimal feature importance ranks (more details are given below)~\cite{covert2020understanding}. Though SHAP was proposed by~\citet{lundberg2017unified} only in 2017, it has already garnered over 2,000 citations. Furthermore, SHAP is being increasingly adopted in the software engineering community to compute feature importance ranks, as evidenced by its usage in recent studies~\cite{viggiato2020trouncing,esteves2020understanding}. Finally, both of these feature importance methods do not require any hyperparameter optimization unlike the CA techniques used by~\citet{jiarpakdee2020empirical} and \citet{peng2020improve}. Appendix \ref{appendix:ca-methods} describes  Permutation and SHAP in more detail.

%% file: 04d-approach.tex
\subsection{Approach}
\label{sec:approach}
\begin{figure*}
    \includegraphics[width=\linewidth,scale=1.5]{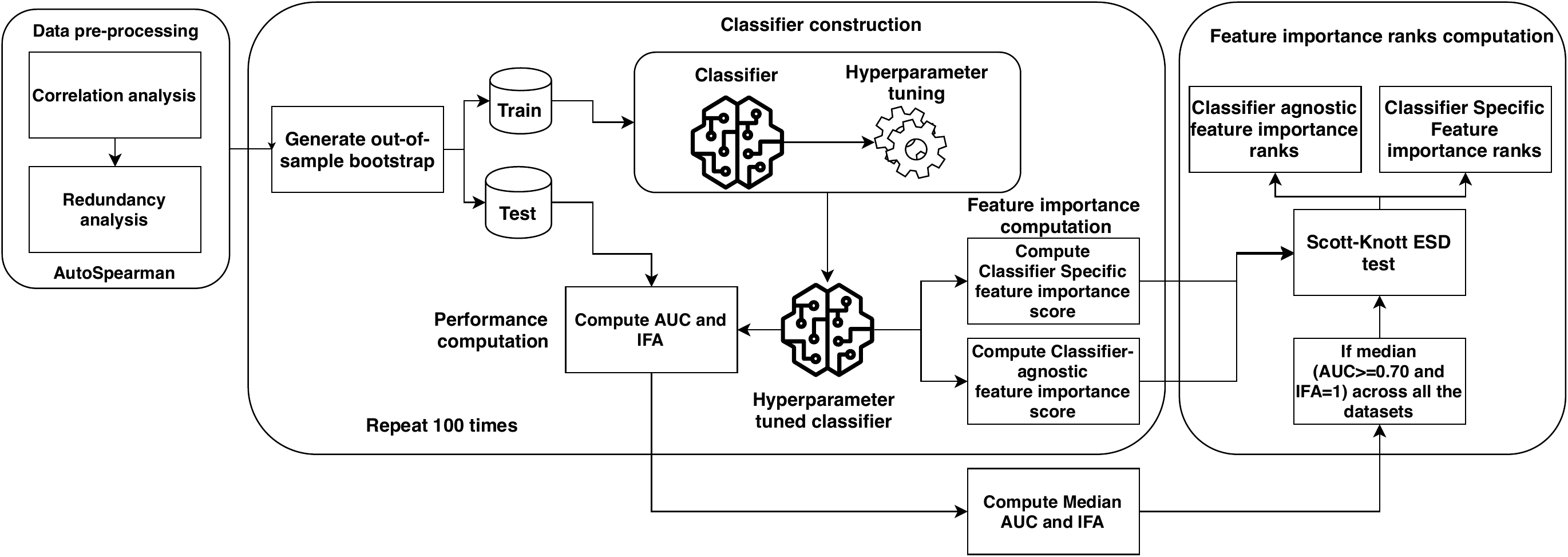}
    \caption{Overview of our case study approach.}
    \label{fig:approach}
\end{figure*}

Figure~\ref{fig:approach} provides an overview of our case study approach. We use this approach to answer all of our aforementioned research questions in Section~\ref{sec:related}.

\subsubsection{Data Pre-processing}

\noindent \textbf{Correlation and redundancy analysis.} We perform correlation and redundancy analysis on the independent features of the studied defect datasets, since the presence of correlated or redundant features impacts the interpretation of a classifier and yields unstable feature importance ranks \citep{jiarpakdee2019impact,tantithamthavorn2018experience,harrell2015regression}. Similar to~\citet{jiarpakdee2020empirical} we employ the AutoSpearman technique~\cite{jiarpakdee2018autospearman} using \texttt{AutoSpearman} function from the \texttt{Rnalytica} R package to remove the correlated and redundant features from our studied datasets.

\subsubsection{Classifier Construction}
\label{subsec:classifier_construction}

\noindent\textbf{Out-of-sample bootstrap.} To ensure the statistical validity and robustness of our findings, we use an out-of-sample bootstrap method with 100 repetitions to construct the classifiers \cite{tantithamthavorn2016empirical,efron1983estimating}. More specifically, for each studied dataset, every classifier is trained 100 times on the 100 resampled \textit{train} sets, then these classifiers are used for computing the 100 feature importance scores. The performance of these trained classifiers are also evaluated on the 100 out-of-sample \textit{test} sets. Appendix \ref{appendix:oosb} describes the out-of-sample bootstrap method in more detail.

\smallskip\noindent\textbf{Classifier construction with hyperparameter tuning.} Several prior studies~\cite{fu2016tuning, tantithamthavorn2018impact} show that hyperparameter tuning is pivotal to ensure that the trained classifiers fit the data well. Furthermore, \citet{tantithamthavorn2018impact} show that feature importance ranks shift between hyperparameter tuned and untuned classifiers. Therefore, we tune the hyperparameters for each of the studied classifiers using random search~\cite{bergstra2012random} in every bootstrap iteration using \texttt{caret} R package~\cite{kuhn2008building}. In every iteration, we pass the \textit{train} set, classifier and the associated hyperparameters outlined in Table~\ref{tab:approach_classifiers} to \texttt{train} function of the \texttt{caret} package. similar to~\citet{jiarpakdee2020empirical}, we then use the automated hyperparameter optimization option provided by train function to conduct a random search (with 10-fold cross-validation) for the optimal hyperparameters that maximizes AUC measure. Once the optimal hyperparameters that maximizes the AUC measure for the given classifier are found, we use these hyperparameters to construct the classifier on the train set. Finally, we use these hyperparameter tuned classifiers to conduct the rest of our study.


\subsubsection{Performance Computation} 
Similar to a recent study by~\citet{jiarpakdee2020empirical}, we compute the AUC (Area Under the Receiver Operator Characteristic Curve) and the IFA (Initial False Alarm) to measure the performance of our classifiers. We choose these two performance measures in particular over other performance measures for the following reasons. First, several prior studies recommend the use of the AUC measure over other performance measures to quantify the discriminative capability of a classifier~\cite{ghotra2015revisiting,lessmann2008benchmarking,rajbahadur2019impact,tantithamthavorn2018experience}. Second, as \citet{parnin2011automated} and \citet{huang2017supervised} argue, the IFA of a classifier being low is extremely important for a classifier to be adopted in practice. For these reasons, we choose the AUC and IFA measures to evaluate the performance of our classifiers. In Appendix \ref{appendix:perf-measures}, we describe provide more details about AUC and IFA, including how they are calculated.



\subsubsection{Computation of Feature Importance Scores} 
We use both the CS and CA methods to computer feature importance scores, as detailed in Section~\ref{sec:fi-methods}, for all the studied classifiers in each bootstrap iteration. For CA methods: we use the \texttt{vip} package and the method outlined by \citet{rajbahadur2017impact} to compute the PDP and Permutation CA methods feature importance scores respectively. For the CS computation, we use the \texttt{VarImp()} function of the \texttt{caret} R package~\cite{kuhn2012variable}.

\subsubsection{Computation of Feature Importance Ranks} 

We use the Scott-Knott Effect Size Difference (SK-ESD) test (v2.0)~\cite{tantithamthavorn2016scottknottesd} to compute the feature importance ranks from the feature importance scores computed in the previous step, as done by prior studies~\cite{rajbahadur2019impact,jiarpakdee2019impact}. For each dataset and studied classifier, three feature importance scores are computed (one CS score and two CA scores) for each bootstrap iteration. The SK-ESD test is applied on these scores to compute three feature importance rank lists (one CS rank list and two CA rank lists) for all the 6 studied classifiers on each dataset. The process of feature importance rank computation from the feature importance scores is depicted in the right-hand side of Figure~\ref{fig:approach}. Also, we note that we only compute the feature importance ranks for classifiers that simultaneously have a median AUC greater than 0.7 and a median IFA = 1. We do so, as \citet{chen2018applications} and \citet{lipton2016mythos} argue, a classifier should have a good operational performance for the computed feature importance ranks to be trusted. Due to this constraint, we discarded the classifier C5.0Rules from our studied classifiers. 


%% file: 04e-metrics.tex
\subsection{Evaluation Metrics}
\label{sec:metrics}

We measure the difference between the different feature importance rank lists by measuring how much they agree with each other.

\smallskip \noindent \textbf{Kendall's Tau coefficient ($\tau$)}~\cite{kendall1948rank} is a widely used non-parametric rank correlation statistic that is used to compute the similarity between \emph{two} rank lists~\cite{kitchenham1995towards,bachmann2010process}. Kendall's~$\tau$ ranges between -1 to 1, where -1 indicates a perfect disagreement and 1 indicates a perfect agreement.
We use the interpretation scheme suggested by \citet{akoglu2018user}: 

\setlength{\abovedisplayskip}{0pt}

\begin{align*}
    \text{Kendall's } \tau \text{ Agreement} = 
        \begin{cases}
            \text{weak,} & \text{if } |\tau| \leq 0.3 \\
            \text{moderate,} & \text{if }  0.3 < |\tau| \leq 0.6 \\
            \text{strong} & \text{if }  |\tau| > 0.6
        \end{cases}
\end{align*}

\noindent \textbf{Kendall's W coefficient} ~\cite{kendall1948rank} is typically used to measure the extent of agreement among \emph{multiple} rank lists given by different raters (CS methods in our case and raters $\ge$ 2). The Kendall's W ranges between 0 to 1, where 1 indicates that all classifiers agree perfectly with each other and 0 indicates perfect disagreement. We use the Kendall's W in RQ3 to estimate extent to which the different feature importance ranks that are computed by CS methods agree across all the studied classifiers for a given dataset. We use the same interpretation scheme for Kendall's W as we use for Kendall's Tau.

\smallskip \noindent \textbf{Top-3 overlap} is a simple metric that computes the amount of overlap that exists between features at the top-3 ranks in relation to the total number of features at the top-3 ranks across n feature importance rank lists. This metric does not consider the ordinality of the features in the top-3 ranks, i.e., the order in which a given feature appears in the top-3 ranks. Rather, it only checks if a given feature appeared in all of the top-3 rank lists. Top-3 overlap is adapted from the popular Jaccard Index~\cite{jaccard1901etude} for measuring similarity. We compute the top-3 overlap among $n$ feature importance lists with the equation~\ref{eqn2} ($k = 3$), where $list_i$ is the $i$th feature list and $n$ is the total number of lists of features to compare (for in RQ1 and RQ2, $n =2$, whereas in RQ3, $n=5$). 

\begin{equation}
    \begin{aligned}
    Top-k~overlap = \frac{\bigcap^n_{i\geq 2}{Features~at~top~k~ranks~of~list_i}}{\bigcup^n_{i\geq 2}{Features~at~top~k~ranks~of~list_i}}
    \end{aligned} \label{eqn2}
\end{equation}

We define the interpretation scheme for Top 3 overlap as follows, which aims to enable easier interpretation of the results:

\begin{align*}
    \text{Top-3 Agreement} = 
        \begin{cases}
            \text{negligible,} & \text{if } 0.00 \leq \text{top-3 overlap} \leq 0.25 \\
            \text{small,} & \text{if }  0.25 < \text{top-3 overlap} \leq 0.50 \\
            \text{medium,} & \text{if }  0.50 < \text{top-3 overlap} \leq 0.75 \\
            \text{large} & \text{if } 0.75 < \text{top-3 overlap} \leq 1.00 
        \end{cases}
\end{align*}

For example, assume that the top-3 features for CS and CA on a given dataset and classifier are $Imp_{CS}(Top~3)= \{cbo,loc,pre\}$ and $Imp_{CA}(Top~3)= \{loc,lcom3,dit\}$ respectively. Then the top-3 overlap corresponds to $1/5=0.2~($as$~n=2, k=3)$.

\smallskip\noindent\textbf{Top-1 overlap} is analogous to the Top-3 overlap metric (Equation~\ref{eqn2}, with $k=1$). We define the interpretation scheme for Top-1 overlap as follows: if top-1 overlap is $\leq$ 0.5 then agreement is low, otherwise agreement is deemed high. 

%% file: 05-casestudy-results.tex
\section{Case Study Results}
\label{sec:rqs}

In this section, we detail the results of our case study with regards to our research questions from Section~\ref{sec:motivation_related}. 

\input{05a-RQ1.tex}
\input{05b-RQ2.tex}
\input{05c-RQ3.tex}

%% file: 05a-RQ1.tex
\subsection{(RQ1) \RQOne}
\label{sec:rq1}


\noindent \textbf{Approach:} For each of the five constructed classifiers with (median AUC $>$ 0.7 and median IFA $\le$ 1 across the studied datasets), we compare the feature importance ranks that are computed by the CA and CS methods across all the studied datasets. For each classifier, on a given dataset, we compare the feature importance ranks computed by SHAP and Permutation CA methods with the feature importance ranks that are computed by the studied CS method of a classifier. We quantify the agreement between the two rank lists in terms of Top-1 overlap, Top-3 overlap and Kendall's Tau. We compute the Top-1 and Top-3 overlap in addition to the Kendall's Tau because some of the prior work primarily examines the top $x$ important features \cite{hassan2005top,lewis2013does,chen2018applications}. Finally, we aggregate the comparisons with respect to each classifier across the studied datasets.

For instance, for the avNNet classifier, we first compare the feature importance ranks that are computed by SHAP (CA method) with those that are computed by the CS method of avNNet (i.e. NNFI, see Table~\ref{tab:CS}) on the \texttt{eclipse-2.0} dataset.  Next, we determine the agreement between the two lists according to Top-1, Top-3 overlap, and Kendall's Tau. We then repeat this step for every dataset and plot the distribution for each agreement metric. An analogous process is followed in order to compare the Permutation method with the NNFI method.

The goal of this RQ is to determine the extent to which the feature importance ranks that are computed by CA methods differ from the more widely used and accepted CS methods for each classifier. If the studied CA methods consistently have a high agreement with the CS methods for each classifier and across all the studied datasets, then one can use both CS methods and CA methods interchangeably.





\begin{figure*}[!htbp]
    \center
    \includegraphics[width=0.75\linewidth]{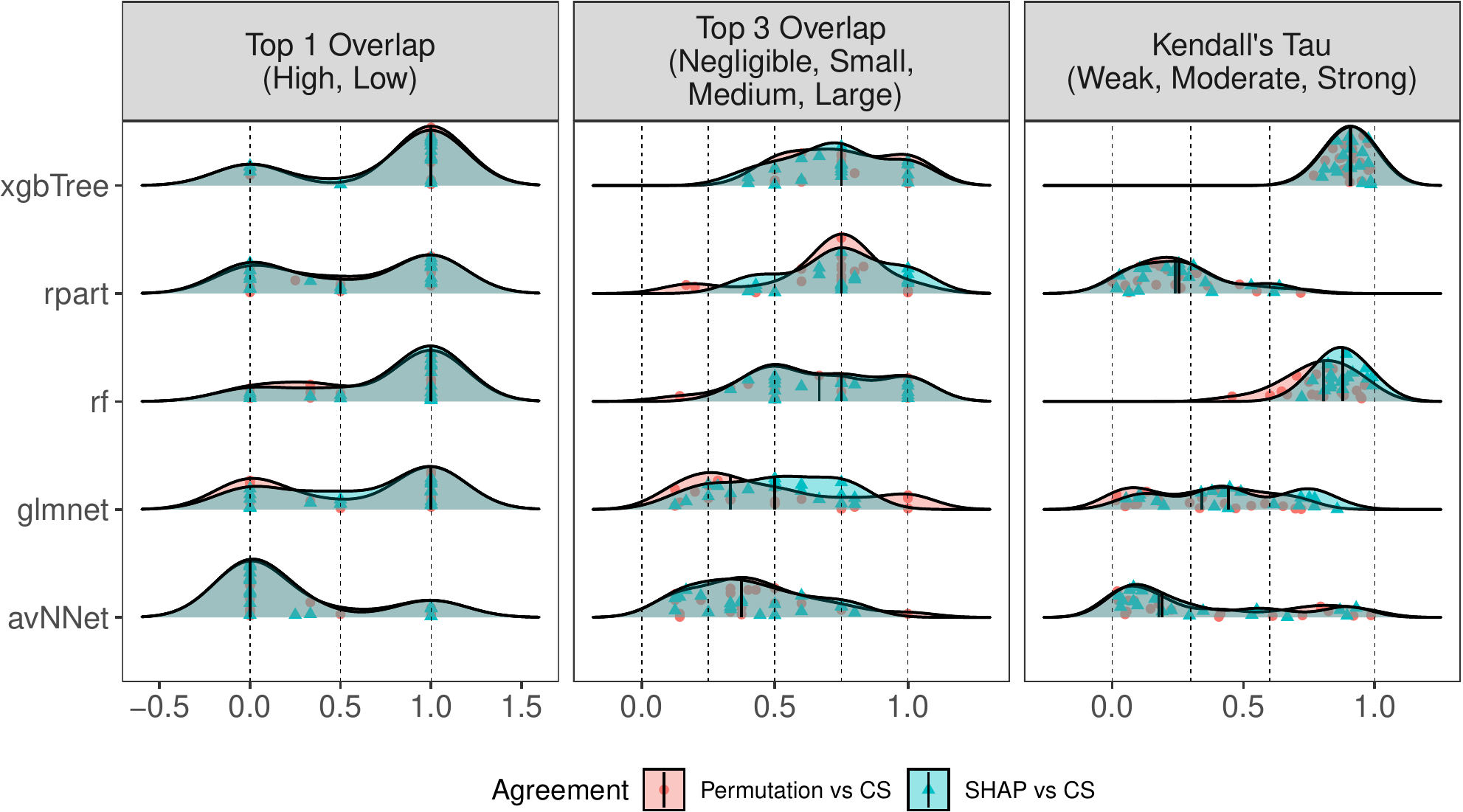}
    \caption{A density plot of Top-1 Overlap, Top-3 Overlap, and Kendall's Tau between CA and CS methods for each classifier across the studied datasets. The circles and triangles correspond to individual observations. The dotted lines correspond to the metric-specific interpretation scheme outlined in Section~\ref{sec:metrics}. The vertical lines inside the density plots correspond to the median of the distributions.}
    \label{fig:rq1_mainplot}
\end{figure*}

\smallskip\noindent\textbf{Results:} \textbf{Result 1) \textit{The SHAP and Permutation CA methods have a low median top-1 overlap with the CS methods on two of the five studied classifiers.}} The leftmost lane in Figure~\ref{fig:rq1_mainplot} shows the top-1 overlap between the feature importance rank lists that are computed by the CS and CA methods for each classifier and across all the studied datasets. We observe that the median top-1 overlap between the studied CA methods and the CS method of a classifier is low for two classifiers, namely rpart and avNNet. In other words, even the most important feature varies between the rankings that are computed by the CA and CS methods for two of the studied classifiers.

\smallskip \noindent \textbf{Result 2) \textit{Both CA methods have a small median top-3 overlap with CS methods on two of the five studied classifiers.}} We see from the middle lane of Figure~\ref{fig:rq1_mainplot} that the features that are reported at the top-3 ranks by both the SHAP and Permutation method do not exhibit a large overlap with the feature importance ranks that are computed by the CS method for \textit{any} of the studied classifiers. Furthermore, from Figure~\ref{fig:rq1_mainplot}, we observe that even on cases where the median overlap is medium, the spread of the density plot is also large (i.e., several datasets exhibit small and even negligible top-3 overlap).


\smallskip \noindent \textbf{Result 3) \textit{For three out of the five studied classifiers, the Kendall's Tau agreement between CA and CS methods is only moderate at best.}} Kendall's Tau values between the feature importance ranks that are computed by CA and CS methods for each classifier and across all the studied datasets are depicted as density distributions in Figure~\ref{fig:rq1_mainplot}. For both rpart and glmnet, the median Kendall's Tau agreement is weak. Even for avNNet, where the median agreement between the CA and CS methods is moderate, the spread of the density lot is large with several datasets exhibiting negligible agreement. We observe that the median Kendall's Tau indicates a strong agreement for only for two of the six studied classifiers, namely xgbTree  and rf (note the vertical bars inside the density plots in the right-most lane). 
In summary, the CA and CS methods do not always exhibit strong agreement for the computed feature importance ranks across the studied classifiers. Therefore, we discourage the interchangeable use of CA and CS methods in general. In particular, we suggest that unless agreement between the CA and CS methods can be improved (we present a potential solution in Section~\ref{sec:disc3}), whenever possible, future defect prediction studies should preferably choose the same feature importance method when replicating or seeking to validate a prior study. 


\smallskip \tastycolorbox{The computed feature importance ranks by CA and CS methods do not always strongly agree with each other. For two of the five studied classifiers, even the most important feature varies across CA and CS methods.}

%% file: 05b-RQ2.tex
\subsection{(RQ2) \RQTwo}
\label{sec:rq2}

\noindent\textbf{Approach:} For each of the studied datasets, we check the extent to which the feature importance ranks computed by SHAP and Permutation CA methods agree with each other for all the five studied classifiers. Similarly to the previous RQ, in order to quantify agreement, we compute the Top-1 Overlap, Top-3 Overlap, and Kendall's Tau measures.





\begin{figure*}[!htbp]
    \center
    \includegraphics[width=0.75\linewidth]{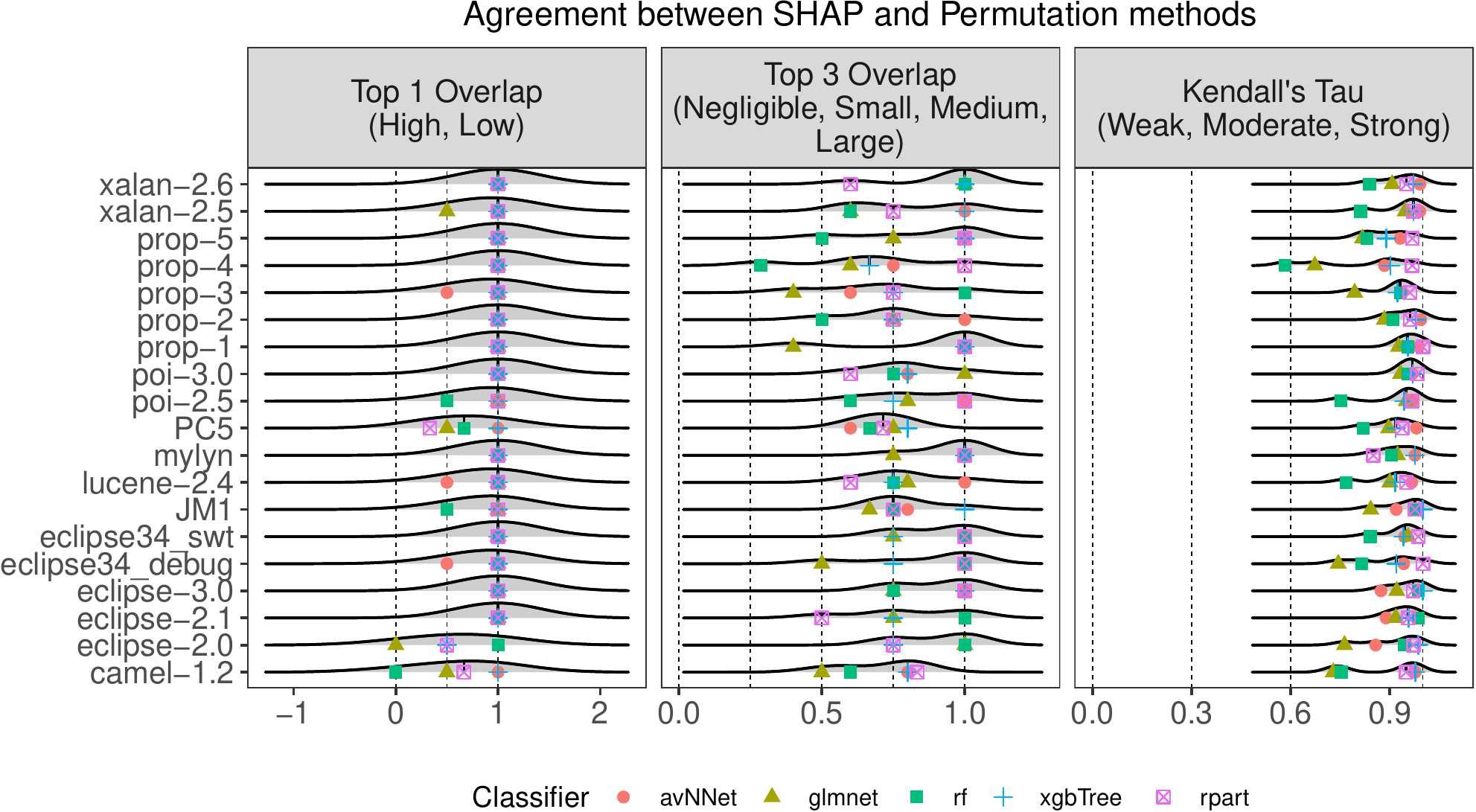}
    \caption{A density plot of Top-1 Overlap, Top-3 Overlap, and Kendall's Tau values between the SHAP and Permutation CA methods for each of the studied classifiers across all the studied datasets. The dotted lines correspond to the metric-specific interpretation scheme outlined in Section 3.5. The vertical lines inside the density plots correspond to the median of the distributions. Please note the empty space in the rightmost figure is due to the observed range of Kendall's Tau values in this case varying between 0.5 and 1. In other words, across all the studied datasets the computed feature importance ranks of SHAP and Permutation at least had a moderate agreement. }
    \label{fig:rq2_mainplot}
\end{figure*}

\smallskip \noindent \textbf{Result 4) \textit{SHAP and Permutation methods have a high median Top-1 overlap and a strong agreement (in terms of Kendall's Tau) across all the studied datasets.}} Furthermore, SHAP and Permutation CA methods do not have a small or negligible overlap across any of the studied datasets. Except on Prop-4 and PC5 datasets, the feature importance ranks computed by SHAP and Permutation CA methods have a large median Top-3 overlap. Furthermore, from the rightmost lane in Figure~\ref{fig:rq2_mainplot} we observe that except for the rf classifier on prop-4 dataset, the Kendall's Tau agreement values are consistently strong for all the studied classifiers. 

\smallskip \tastycolorbox{The computed feature importance ranks by the studied CA methods exhibit a strong agreement including the features reported at top-1 and top-3 ranks for a given dataset and classifier.}

%% file: 05c-RQ3.tex
\subsection{(RQ3) \RQThree}
\label{sec:rq3}

\noindent\textbf{Approach:} For each of the datasets, we obtain the computed feature importance ranks by the studied CS methods of each of the five studied classifiers. We then calculate the Kendall's W between the six feature importance rank lists that are computed by the studied CS method of each classifier. Unlike the previous RQs, we compute the Kendall's W instead of Kendall's Tau, as Kendall's W is able to measure agreement among multiple feature importance rank lists (Section~\ref{sec:metrics}). Furthermore, we also calculate the Top-3 overlap among all the five feature importance rank lists. We do so for all the studied datasets. A high Kendall's W and a high Top-3 overlap across all the studied datasets among the constructed classifiers would indicate high agreement between the computed feature importance ranks  by different classifiers and the studied CS method of each classifier.

\smallskip\noindent\textbf{Results:} \textbf{Result 5) \textit{The computed feature importance ranks by different CS methods vary extensively.}} None of CS methods agree on the most important feature (as evidenced by the results presented in Table~\ref{tab:RQ3_table}). Furthermore, the maximum top-3 overlap is only small and it happens for only three out of 18 datasets. Finally, we also observe that, on a given dataset, only on two occasions the feature importance rank lists computed by the different CS methods strongly agree with each other. We summarize the Top 1 overlap, the Top-3 overlap and Kendall's W among the computed feature importance ranks for all the six studied classifiers across the studied datasets in Table~\ref{tab:RQ3_table}.

\begin{table}[!htbp]
  \centering
  \caption{Top-1 overlap, Top-3 overlap, and Kendall's W among the computed feature importance ranks by the CS method of each classifier. Best results for each metric are shown in bold.}
  \begin{threeparttable}
    \begin{tabular}{l|p{1.7cm}|p{1.7cm}|p{1.5cm}}
    \hline
    \textbf{Dataset} & \textbf{Top-1 overlap}& \textbf{Top-3 overlap} &\textbf{Kendall's W}  \\
    \hline
    poi-3.0 & \multicolumn{1}{r|}{Low (0)} & \multicolumn{1}{r|}{Negligible (0)} & \multicolumn{1}{r}{Weak (0.13)} \\
    camel-1.2 & \multicolumn{1}{r|}{Low (0)} & \multicolumn{1}{r|}{Negligible (0)} & \multicolumn{1}{r}{Weak (0.22)} \\
    xalan-2.5 & \multicolumn{1}{r|}{Low (0)} & \multicolumn{1}{r|}{Negligible (0.10)}&  \multicolumn{1}{r}{Weak (0.18)} \\
    xalan-2.6 & \multicolumn{1}{r|}{Low (0)} & \multicolumn{1}{r|}{Negligible (0.16)}&  \multicolumn{1}{r}{Weak (0.25)} \\
    eclipse34\textunderscore debug & \multicolumn{1}{r|}{Low (0)}& \multicolumn{1}{r|}{Negligible (0.14)}& \multicolumn{1}{r}{Weak (0.26)} \\
    eclipse34\textunderscore swt & \multicolumn{1}{r|}{Low (0)} & \multicolumn{1}{r|}{small (0.33)} & \multicolumn{1}{r}{\textbf{Strong (0.62)}} \\
    pde & \multicolumn{1}{r|}{Low (0)} & \multicolumn{1}{r|}{\textbf{Small (0.4)}} & \multicolumn{1}{r}{\textbf{Strong (0.70)}}\\
    PC5 &  \multicolumn{1}{r|}{Low (0)} & \multicolumn{1}{r|}{Negligible (0)} & \multicolumn{1}{r}{Weak (0.18)}\\
    mylyn & \multicolumn{1}{r|}{Low (0)} & \multicolumn{1}{r|}{\textbf{Small (0.29)}} & \multicolumn{1}{r}{Weak (0.16)}\\
    eclipse-2.0 & \multicolumn{1}{r|}{Low (0)} & \multicolumn{1}{r|}{Negligible (0.17)} & \multicolumn{1}{r}{Weak (0.26)}\\
    JM1 & \multicolumn{1}{r|}{Low (0)} & \multicolumn{1}{r|}{Negligible (0.20)} & \multicolumn{1}{r}{\textbf{Moderate (0.31)}} \\
    eclipse-2.1& \multicolumn{1}{r|}{Low (0)} & \multicolumn{1}{r|}{Negligible (0.17)} & \multicolumn{1}{r}{\textbf{Moderate (0.31)}}\\
    prop-5 & \multicolumn{1}{r|}{Low (0)} & \multicolumn{1}{r|}{Negligible (0)} & \multicolumn{1}{r}{Weak (0.17)}\\
    prop-4 & \multicolumn{1}{r|}{Low (0)} & \multicolumn{1}{r|}{Negligible (0.14)} & \multicolumn{1}{r}{\textbf{Moderate (0.37)}}\\
    prop-3 & \multicolumn{1}{r|}{Low (0)} & \multicolumn{1}{r|}{Negligible (0.09)} & \multicolumn{1}{r}{\textbf{Moderate (0.32)}}\\
    eclipse-3.0 & \multicolumn{1}{r|}{Low (0)}  & \multicolumn{1}{r|}{\textbf{Small (0.28)}} & \multicolumn{1}{r}{\textbf{Moderate (0.32)}}\\
    prop-1 & \multicolumn{1}{r|}{Low (0)} & \multicolumn{1}{r|}{Negligible (0.16)} & \multicolumn{1}{r}{\textbf{Moderate (0.36)}}\\
    prop-2 & \multicolumn{1}{r|}{Low (0)} & \multicolumn{1}{r|}{Negligible (0)}& \multicolumn{1}{r}{Weak (0.28)}\\
    \hline
    \end{tabular}%

\end{threeparttable}
  \label{tab:RQ3_table}%
\end{table}%

From Table~\ref{tab:RQ3_table}, we observe that both the Kendall's W and top-3 overlap among the feature importance ranks that are computed by studied CS methods associated with each of the classifier -- which are widely used in the software engineering community-- is very low. Such a small Top-3 overlap and Top-1 overlap for all the datasets indicates that computed feature importance ranks by CS methods differ substantially among themselves. Hence, different classifiers and their associated CS methods cannot be used interchangeably. 


\smallskip \tastycolorbox{On a given dataset, even the commonly used CS methods yield vastly different feature importance ranks, including the top-3 and the top-1 most important feature(s).}

%% file: 06-discussion.tex
\section{Discussion}
\label{sec:disc}

\input{06b-disc2.tex}

\input{06c-disc3.tex}

%% file: 06b-disc2.tex
\subsection{Why do different feature importance methods produce different top-3 features on a given dataset?}
\label{sec:disc2}

\textbf{Motivation:} From the results presented for RQ1 (Section~\ref{sec:rq1}) we observe that, on a given dataset and classifier the CA methods and CS methods produce different feature importance ranks (including the top-3 ones). Similarly, from the results presented in RQ3 (Section~\ref{sec:rq3}), on a given dataset, even the widely used CS methods produce vastly different feature importance ranks. Such a result is in spite of us having removed the correlated and redundant features from the datasets in a pre-processing step using a state-of-the-art technique like AutoSpearman~\cite{jiarpakdee2018autospearman}. However, in contrast to the results presented in RQ1 and RQ3, in RQ2 (Section~\ref{sec:rq2}), we observe that the studied CA methods (i.e., Permutation and SHAP methods) produce similar feature importance ranks on a given dataset and classifier.




We hypothesize that different CS methods produce different top-3 feature importance ranks even on the same dataset and classifier (and when compared to the feature importance ranks computed by the CA methods) because of feature interactions that are present in the studied datasets. Feature interactions can be defined as a phenomenon where the effect of the independent features on a dependent feature is not purely additive~\cite{freeman1985analysis,molnar2018interpretable}). We arrive at such a hypothesis as many of the prior studies show that the presence of feature interactions in a given dataset can affect the different feature importance methods differently and make them assign different feature importance ranks to features~\cite{de2007interpretation, freeman1985analysis, fisher2018all,devlin2019disentangled,lundberg2018consistent}. 
Therefore, in this section, we seek to find out if different feature importance methods (in addition to being inherently different) yield different feature importance ranks due to the presence of feature interactions in the dataset.

We further note that computed feature importance ranks by the CA methods exhibit a high agreement and overlap as both Permutation and SHAP are typically not impacted by the feature interactions present in a dataset~\cite{molnar2018interpretable}.

\noindent \textbf{Approach:} To test our hypothesis and detect if any of the features presented in a given dataset interact with other features in that dataset, we compute the \textit{Friedman H-Statistic}~\cite{friedman2008predictive} for each feature against all other features. The Friedman H-statistic works as follows. First, a classifier (any classifier - we use random forest as it captures interactions well~\cite{wright2016little}) is constructed using the given dataset. For instance, consider the Eclipse-2.0 dataset and that we wish to compute the Friedman H-Statistic between the feature \texttt{pre} and all the other features. We first compute the partial dependence between \texttt{pre} and the dependent variable with respect to the random forest classifier (\texttt{PD\_pre}) for all the data points. Following which, partial dependence between all the features (as a single block) in Eclipse-2.0 (except \texttt{pre}) is computed (\texttt{PD\_rest}) for all the data points. If there is no feature interaction between \texttt{pre} and the other features in Eclipse-2.0, the outcome probability of the constructed classifier can be expressed as a sum of \texttt{PD\_pre} and \texttt{PD\_rest}. Therefore, the difference between the outcome probability of the classifier and the sum of \texttt{PD\_pre} and \texttt{PD\_rest} is computed and the variance of this difference is reported as the Friedman H-Statistic. We compute the Friedman H-Statistic for all the studied datasets 10 times, as Friedman H-Statistic is known to exhibit fluctuations because of the internal data sampling~\cite{molnar2018interpretable}. We then consider the median score of the Friedman H-Statistic for each dataset. We use the R package~\texttt{iml}\footnote{\url{https://cran.r-project.org/web/packages/iml/index.html}} to compute the Friedman H-Statistic.

The Friedman H-Statistic is a numeric score that ranges between 0 and 1. However, it can sometimes exceed 1 when the higher order interactions are stronger than the individual features~\cite{molnar2018interpretable}). A Friedman H-Statistic of 0 or closer to 0 indicates that no interaction exists between the given feature and the rest of the features and a Friedman H-Statistic of 1 indicates extremely high levels of interaction. For a more theoretical and detailed explanation we refer the readers to~\cite{friedman2008predictive, molnar2018interpretable}. In this study, we consider a feature to exhibit interactions with other features if the Friedman H-Statistic is $\geq$ 0.3. We choose 0.3 as a cut-off, to only indicate the existence of a feature interaction, but not to qualify the strength of the interaction, because the presence of feature interactions irrespective of the strength could potentially impact feature importance ranks~\cite{de2007interpretation, freeman1985analysis, fisher2018all,devlin2019disentangled,lundberg2018consistent}. In addition we also report the results for the number of features that exhibit a Friedman H-statistic $\geq$ 0.5. We choose to report the results on multiple thresholds to present a comprehensive depiction of the feature interactions. Furthermore, there is no established guideline in the literature regarding how one should interpret the Friedman H-statistic or how thresholds should be selected.

\smallskip\noindent\textbf{Construction of a synthetic dataset without any feature interactions.} Next, to determine if the absence of feature interactions enables the different CS methods to compute the same top-3 features, we simulate a dataset with no feature interactions. We do so instead of using a real dataset as it is difficult to find a real-world defect dataset without any feature interactions. We generate a dataset with 1,500 data points and 11 independent features of which five features carry the signal: \texttt{signal~=~\{x1,x2,x3,x4,x5\}} and six features are just noise i.e., does not exhibit any relationship to the dependent feature \texttt{noise~=~\{n1, n2, n3, n4, n5, n6\}}. We add noise features to make our simulated dataset similar to that of a real-world defect dataset. All the signal features and \texttt{n1, n5, n6} are generated by randomly sampling the normal distribution with mean~=~0 and standard deviation~=~1. Similarly, we sample the uniform distribution between 0 and 1 to generate the values for \texttt{n2, n3, n4}. We use both the normal and the uniform distributions for generating the noise features to ensure the presence of different types of noise in our simulated dataset. Next, to construct our dependent feature for the dataset, we construct $y_{signal}$ with the signal variables as given in Equation~\ref{disc2:eqn1}. We assign different weights to the different signal features when constructing the $y_{signal}$ to ensure that we know the true importance of each of the signal features. We then convert the $y_{signal}$ in to a probability vector $y_{prob}$ with a sigmoid function as given in Equation~\ref{disc2:eqn2}. Finally, we generate the dependant feature by sampling the binomial distribution to generate the dependent feature $y_{dependant}$ with $y_{prob}$ as given in Equation~\ref{disc2:eqn3}.

\begin{equation}
    y_{signal} = 20x1 + 10x2 + 5x3 + 2.5x4 + 0.5x5 \\
    \label{disc2:eqn1}
\end{equation}
\begin{equation}
        y_{prob} =  \frac{\mathrm{1} }{\mathrm{1} + e^{-y_{signal}}} \\
   \label{disc2:eqn2}
\end{equation}
\begin{equation}
     y_{dependant} = Binomial(1500,y_{prob})
     \label{disc2:eqn3}
\end{equation}

\smallskip\noindent\textbf{Construction of classifiers and calculation of top-1 and top-3 overlaps.} We then construct all of the studied classifiers on the simulated dataset with $y_{dependent}$ as the dependent feature. We construct all the classifiers with 100-Out-of-sample bootstrap on the simulated dataset and compute the feature importance ranks computed by the CA and CS methods as outlined in Section~\ref{sec:approach}. For each of the studied classifiers, we calculate the top-1 and top-3 overlap between the feature importance ranks computed by the CA and CS methods (similarly to RQ1). Furthermore, we also calculate the top-1 and top-3 overlaps between feature importance ranks computed by the studied CS methods (similarly to RQ3). We then check if they exhibit a top-1 and top-3 overlap close to 1 for all the classifiers between the computed feature importance ranks of the CS and the CA methods. In addition we also check the top-1 and top-3 overlap among the computed feature importance ranks of the various CS methods. If in both the cases they exhibit a top-1 and top-3 overlap close to 1, we can then assert that the feature interactions in the dataset affects the top-3 features computed by the different CS methods and vice versa.

\smallskip\noindent\textbf{Determining whether feature interactions impact interpretation.} To verify that the feature interactions present in the dataset impact only the studied CS methods and not the studied CA methods, we compute the feature importance ranks for all the studied classifiers with the studied CS and CA methods on a simulated dataset with feature interactions. To simulate a dataset with interactions, we take the simulated dataset from earlier (the one without any interactions) and introduce interactions by modifying the Equation~\ref{disc2:eqn1}. We modify the $y_{signal\_with\_interactions}$ to depend on hidden interactions as given by Equation~\ref{disc2:eqn4}. The rest of the data generation process remains the same as earlier i.e., the independent features are still given by \texttt{independent features~=~\{signal,noise\}} where  \texttt{signal~=~\{x1,x2,x3,x4,x5\}} and \texttt{noise~=~\{n1, n2, n3, n4, n5, n6\}}. Finally, $y_{dependant\_with\_interactions}$ is generated using Equations~\ref{disc2:eqn2} and~\ref{disc2:eqn3}.

\begin{equation}
    y_{signal\_with\_interactions} = y\_{signal} + x1*x3 + x2*x3 + x2*x1 \\
    \label{disc2:eqn4}
\end{equation}

We then construct all of the studied classifiers with $y_{dependent\_with\_interactions}$ on the simulated dataset with interactions using 100-Out-of-sample bootstrap. Next, we compute the feature importance ranks using the studied CA and CS methods. Finally, we calculate the top-1 and top-3 overlap between the computed feature importance ranks of CA and CS methods, respectively (similarly to RQ2 and RQ3). If feature interactions do not impact the CA methods, we should observe high top-1 and top-3 overlaps (close to 1) between the computed feature importance ranks of the CA methods and vice versa. Similarly, if the feature interactions impact the computed feature importance ranks of CS methods, we should observe low top-1 and top-3 overlaps between the computed feature importance ranks of the different CS methods and vice versa.

\smallskip \noindent \textbf{Results: Result 6)} \textbf{\textit{ At least two features and as many as eight features interact with the rest of the features in all of the 18 studied datasets (i.e., Friedman H-Statistic $\geq 0.3$). Furthermore, we find that 14 of the 18 datasets have at least two features with a Friedman H-Statistic $\geq$ 0.5.}} We present the number of features in each dataset with a Friedman H-statistic $\geq$ 0.3 and $\geq$ 0.5 in Table~\ref{tab:disc2_tab1} (Appendix~\ref{appendix:friedman}). From Table~\ref{tab:disc2_tab1}, we observe that all datasets contain more than two features that interact with the other features. Though Friedman H-Statistic only computes if a given feature interacts with the rest of the features and excludes other feature interactions like second-order interactions, pairwise interactions, and higher-order interactions, it gives us a hint as to the presence or absence of feature interactions in a dataset.

\smallskip \noindent \textbf{Result 7)} \textbf{\textit{The top-3 and the top-1 overlap between the feature importance ranks computed by the CS and CA methods on each of the classifiers is 1 on our simulated dataset devoid of feature interactions.}} In addition, the top-3 and top-1 overlap between the computed feature importance ranks of the CS methods on the simulated dataset without interaction is 1. Such a result indicates that in the dataset without feature interactions, all the studied feature importance methods identify the same top-3 features. Furthermore, we also observe that all the studied important features are identified {x1, x2, x3} as the top-3 features in the same order of importance. Thus, we assert that the different feature importance methods are able to assign the feature importance ranks correctly when independent features' contribution to the dependent feature is additive without any interactions.

\smallskip \noindent \textbf{Result 8)} \textbf{\textit{The top-3 and the top-1 overlap between the feature importance ranks computed by the studied CA methods is 1 on the simulated dataset with interactions.}} In turn, we find that the top-1 and top-3 overlap between the feature importance ranks computed by the different CS methods is 0 on the simulated dataset with interactions. Such a result indicates that studied CA methods (i.e., SHAP and Permutation) are not impacted by the feature interactions in the simulated dataset. However, the computed feature importance ranks of the studied CS methods are heavily influenced by the presence of feature interactions in the dataset. 

Hence, we conclude that the presence of feature interactions in the studied defect datasets could be the reason why different CS methods produce a different top-3 set of features. In addition, we conclude that alongside the fact that different feature importance methods compute feature importances differently, feature interactions in the datasets can also be a key confounder that affects the computed feature importance ranks by the studied feature importance methods.



%% file: 06c-disc3.tex
\subsection{Can we mitigate the impact of feature interactions?}
\label{sec:disc3}

\textbf{Motivation:} From the results presented in Section~\ref{sec:disc2} we observe that feature interactions impact feature importance ranks computed by the CS methods. Such a result indicates that CS methods cannot be interchangeably used. Though comprehensively identifying and removing all the feature interactions from a dataset is still an open area of research, there exist several simple methods like Correlation-based Feature Selection (CFS)~\cite{hall1999correlation}, wrapper methods~\cite{hall1999correlation} that allows us to remove lower order feature interactions~\cite{arisholm2010systematic,cahill2013predicting,d2012evaluating}. 

Therefore, in this discussion, we investigate if removing the feature interactions present in a dataset through a simple method like CFS would increase the agreement between the computed feature importance ranks of the studied CS methods on a given dataset. We also investigate if removing feature interactions yields an improved agreement between the computed feature importance ranks of CA and CS methods. Finally we also study if the removal of feature interactions has any impact on the agreement between the feature importance ranks computed by the studied CA methods. If it does, then we can recommend researchers and practitioners to remove the feature interactions in their datasets using CFS prior to building the machine learning classifiers.

\smallskip\noindent\textbf{Approach:} We use the CFS method to remove the feature interactions across the studied datasets. We use the CFS method in lieu of other methods like Wrapper in the dataset for the following reasons. First, several prior studies show that, in addition to eliminating correlation between the features, CFS method is also useful for mitigating feature interactions~\cite{hall1999correlation,hall2003benchmarking}. Second, CFS method has been widely used in the software engineering community (though for removing correlated features)~\cite{jiarpakdee2020empirical}. Third, it is simple to implement and is extremely fast. Finally, since its a filter type method, it does not make any assumptions about the dataset or the subsequent classifier that is to be built.

Therefore, for each of the studied datasets, following the removal of correlated and redundant features using AutoSpearman, we apply the CFS~\cite{hall1999correlation} method to remove the feature interactions. The CFS method chooses a subset of features that exhibits the strongest relationship with the dependent feature while exhibiting a minimal correlation among themselves. It is important to note that we apply the CFS method to the features that were not flagged as correlated/redundant by AutoSpearman technique (as opposed to applying CFS method by itself to eliminate both the observed inter-feature correlation and feature interaction), as~\citet{jiarpakdee2020empirical} argue CFS method might not remove all the correlated features from the dataset effectively. 

After removing the feature interactions using CFS methods, we re-run the analysis that we conducted in RQ1 (Section~\ref{sec:rq1}) and RQ3 (Section~\ref{sec:rq3}) using the same approach. We then evaluate whether the agreement in terms of top-1 and top-3 overlap increases between the CA and CS methods (compared to the results presented in Section~\ref{sec:rq1}). Similarly, we also evaluate whether the agreement between the computed feature importance ranks of different CS methods on a given dataset increases in terms of top-1 and top-3 overlaps (compared to the results presented in Section~\ref{sec:rq3}).

\begin{table}[htbp]
  \centering
  \caption{The median top-1 and top-3 overlap improvements upon removal of feature interactions (FI) between the computed feature importance ranks of the studied CA and CS methods.}
    \begin{tabular}{l|p{1.4cm}|p{1.4cm}|p{1.4cm}|p{1.4cm}}
    \hline
    \multicolumn{1}{c|}{\multirow{2}[4]{*}{\textbf{Classifier}}} & \multicolumn{2}{c|}{\textbf{Top-1 Overlap}} & \multicolumn{2}{c}{\textbf{Top-3 Overlap}} \\
\cline{2-5}    \multicolumn{1}{c|}{} & \textbf{RQ1 results} & \textbf{After removal of FI} & \textbf{RQ1 results} & \textbf{After removal of FI} \bigstrut\\
    \hline
    xgbTree & High & High & Medium & \textbf{Strong} \\
    rpart & Low & \textbf{High} & Medium & \textbf{Strong} \\
    rf & High & High & Medium & \textbf{Strong} \\
    glmnet & High & High & Small & \textbf{Moderate} \\
    avNNet & Low & Low & Small & \textbf{Moderate} \\
    \hline
    \end{tabular}%
  \label{tab:disc3_tab1}%
\end{table}%

\begin{table}[htbp]
  \centering
  \caption{The median top-1 and top-3 overlap improvements upon removal of feature interactions (FI) between the computed feature importance ranks of the studied CS methods.}
    \begin{tabular}{l|p{1.2cm}|p{1.2cm}|p{1.2cm}|p{1.2cm}}
    \hline
    \multicolumn{1}{c|}{\multirow{2}[4]{*}{\textbf{Dataset}}} & \multicolumn{2}{c|}{\textbf{Top-1 Overlap}} & \multicolumn{2}{c}{\textbf{Top-3 Overlap}} \\
\cline{2-5}    \multicolumn{1}{c|}{} & \textbf{RQ3 results} & \textbf{After removal of FI} & \textbf{RQ3 results} & \textbf{After removal of FI} \\
    \hline
    poi-3.0 & Low & Low & Negligible & Negligible \\
    camel-1.2 & Low & Low & Negligible & \textbf{Small} \\
    xalan-2.5 & Low & Low & Negligible & \textbf{Medium} \\
    xalan-2.6 & Low & \textbf{High} & Small & Small \\
    eclipse34\_debug & Low & \textbf{High} & Small & Small \\
    eclipse3\_swt & Low & Low & Negligible & \textbf{Small} \\
    pde & Low & Low & Small & Small \\
    PC5 & Low & Low & Negligible & Negligible \\
    mylyn & Low & Low & Small & Small \\
    eclipse-2.0 & Low & Low & Negligible & Negligible \\
    JM1 & Low & Low & Negligible & \textbf{Small} \\
    eclipse-2.1 & Low & Low & Negligible & \textbf{Small} \\
    prop-5 & Low & \textbf{High} & Negligible & Negligible \\
    prop-4 & Low & \textbf{High} & Negligible & \textbf{Small} \\
    prop-3 & Low & Low & Negligible & Negligible \\
    eclipse-3.0 & Low & Low & Small & Small \\
    prop-1 & Low & Low & Negligible & \textbf{Small} \\
    prop-2 & Low & Low & Negligible & \textbf{Small} \\
    \hline
    \end{tabular}%
  \label{tab:disc3_tab2}%
\end{table}%

\smallskip\noindent\textbf{Results: Result 9) \textit{Removing the feature interactions increases the median top-1 and top-3 overlap between the studied CA and CS methods across all the five studied classifiers.}} From Table~\ref{tab:disc3_tab1} we observe that across all the classifiers, the top-1 and top-3 overlaps improve by at least one level. Such a result indicates that, removing feature interactions with CFS generally yields a higher agreement between the CA and CS methods with regards to feature ranking (i.e., promotes stability of the results).

\smallskip\noindent\textbf{Result 10) \textit{Removing the feature interactions increases the median top-1 and top-3 overlap between the studied CS methods on four and five of the 18 studied datasets respectively.}} Table~\ref{tab:disc3_tab2} depicts the improvements in top-1 and top-3 overlaps between the studied CS methods across all the studied datasets. From Table~\ref{tab:disc3_tab2} we observe that the removal of feature interactions with CFS yields only small improvements. Therefore, we suggest that researchers and practitioners should be cautious when using different CS methods interchangeably even after removing feature interactions. However, our inference is exploratory in nature and thus further research should be conducted to understand advanced feature interaction removal methods may help improve the agreement across different CS methods.

\smallskip\noindent\textbf{Result 11)\textit{ After removing the feature interactions, SHAP and Permutation yield a strong agreement on all datasets.}} From the results presented in Section 4.2 we observe that the Permutation and SHAP had less than large overlap only on the Prop-4 and PC5 datasets. However, upon removal of feature interactions from these two datasets, the feature importance ranks computed by the studied CA methods have a large top-1 and top-3 overlap across all the studied datasets and classifiers.

%% file: 07-implications.tex
\section{Implications}
\label{sec:implications}

In this section, we outline the implications that one can derive from our results, including potential pitfalls to avoid and future research opportunities.

\smallskip\noindent\textbf{Implication 1)~\textit{Researchers and practitioners should be aware that feature interactions can hinder classifier interpretation. We recommend these stakeholders to detect feature interactions in their datasets (e.g., by means of the Friedman's H-statistic) and, in the positive scenario, remove these interactions if possible (e.g., by preprocessing the dataset with the CFS method)}}. From Section~\ref{sec:disc2} and Section~\ref{sec:disc3}, we find that removing the feature interactions, even with a simple method like CFS, increases the agreement between the feature importance rankings produced by the studied feature importance methods. In other words, once feature interactions are removed, the final feature ranking tends to change. Hence, we consider that feature interactions hinder the interpretability of machine learning models in defect prediction. We thus encourage researchers and practitioners to detect feature interactions in their dataset (e.g., by means of the Friedman's H-statistic). In case interactions are discovered, we encourage these stakeholders to remove them if possible (e.g., by preprocessing the dataset with the CFS method).

\smallskip \noindent \textbf{Implication 2)~\textit{The lack of clear specification of the feature importance method employed in software engineering studies seriously threatens the reproducibility of these studies and the generalizability of their insights.}} 14\% of the studies listed in Table~\ref{tab:motivation} do not specify their employed feature importance method to arrive at their insights. 
The absence of the specification of the feature importance method employed is more prevalent for random forest classifiers (3/11 studies) -- the classifier that is widely used in software engineering. This poses a serious threat, as many random forest implementation across the popular data mining toolboxes come with many different ways of computing the feature importance. For instance, random forest implementation in the R package \textit{randomForest}\footnote{\url{https://cran.r-project.org/web/packages/randomForest/index.html}} has 3 feature importance methods available and the R package~\textit{partykit}\footnote{\url{https://cran.r-project.org/web/packages/partykit/index.html}} has 2 implementations of feature importance methods for random forest.

\smallskip\noindent\textbf{Implication 3)~\textit{Future research should evaluate the extent to which SHAP and Permutation can be used interchangeably.}} We conjecture that using either SHAP or Permutation would lead to similar rankings for defect datasets that have similar characteristics to those studied by us. We also conjecture that the feature importance rankings of prior studies in defect classification would not change much if Permutation were to be replaced with SHAP or vice-versa. Nonetheless, we do emphasize that even small changes in rankings could be meaningful in practice and lead to completely different action plans (e.g., in terms of prioritizing pieces of code to be tested or reviewed). Hence, despite the similar rankings produced by SHAP and Permutation, concluding that they can be used interchangeably would be an overstatement. We thus invite future work to further evaluate the differences in the rankings produced by SHAP and Permutation (e.g., by evaluating defect datasets that have different characteristics compared to the ones that we studied).

\smallskip\noindent\textbf{Implication 4)~\textit{Future research should consider investigating whether more advanced feature interaction removal methods can increase the agreement between the feature interaction rankings produced by different feature importance methods.}} By means of a simple feature interaction removal method like CFS, we are able to improve the agreement between the feature importance rankings produced by CS and CA methods (Section~\ref{sec:disc3}). Even between the computed feature importance ranks of CS methods, removing feature interactions helps to improve agreement in some cases. Such a result suggests that more sophisticated feature interaction removal methods have the potential to further improve the agreement between the computed feature importance ranks of different feature importance methods.

%% file: 08-threats.tex
\section{Threats to Validity}
\label{sec:threat}

In the following, we discuss the threats to the validity of our study.

\smallskip \noindent \textbf{Internal validity.} We choose classifier families who has a CS method. As previous studies show that different classifiers may have different performance on a given dataset~\cite{rajbahadur2017impact,ghotra2015revisiting}, this could be a potential threat. However, we choose representative classifiers from 6 of the 8 commonly used classifier families as outlined by \citet{ghotra2015revisiting}.

We use the AUC and IFA performance measures to shortlist the classifiers used in this study. However, using different performance measures like P\textsubscript{opt20}, MCC, and F-Measure to shortlist the classifiers to include in our study might potentially bias the conclusion of our study and we declare it as a threat to internal validity. We invite the future studies to revisit our study by using different performance measures to shortlist the classifiers to be used to compare feature importance ranks computed by different feature importance methods.

The classifiers that we choose to study are either probabilistic or stochastic in nature. We do not include any deterministic classifiers like Naive Bayes in our study and it could be a potential threat to the internal validity of our study. Particularly since,~\citet{wu2008top} point out that computed feature importance ranks of classifiers like decision trees can be particularly sensitive to data characteristics. We invite the future research to revisit our findings on deterministic and stable learners. 

\smallskip \noindent \textbf{Construct validity.} In our study, we choose classifiers where all the studied datasets achieved an AUC above 0.70. According to \citet{muller2005can}, an AUC score above 0.70 indicates the fair discriminative capability of a classifier. Furthermore, these datasets have been used in many of the studies as outlined in Table~\ref{tab:motivation}.

We use a simple random search method to hyperparameter tune our studied classifiers. Our decision stems from the work of~\citet{tantithamthavorn2018impact1}  who show that, irrespective of the performance measure considered, different hyperparameter tuning methods (including grid search, random search, differential evolution based methods and genetic algorithm based methods) yield similar performance improvements. The authors suggest that, as far as performance improvements are concerned, researchers and practitioners can safely use any of the aforementioned automated hyperparameter tuning methods to tune defect classifiers. Therefore, we consider that employing different or more advanced hyperparameter optimization methods would not necessarily result in better hyperparameters for our studied classifiers. Nonetheless, we invite future work to revisit our findings by using more advanced hyperparameter tuning methods to tune the classifiers.

The hyperparameters that we tune for our studied classifiers are limited. Similar to~\citet{jiarpakdee2020empirical}, we use the automated hyperparameter optimization option provided by caret package to tune the hyperparameters of the studied classifiers. caret package only supports tuning a limited number of hyperparameters (please see Table~\ref{tab:approach_classifiers}). Tuning a wider range of hyperparameters for the studied classifiers might potentially change our findings. We consider it a threat and invite the future studies to revisit our findings by tuning a wider range of hyperparameters for the studied classifiers.

\smallskip \noindent \textbf{External validity.} In this study, we choose 18 datasets that represent software projects across several corpora (e.g., NASA and PROMISE) and domains (both proprietary and open-source). However, our results likely do not generalize to all software defect datasets. Nevertheless, the datasets that we use in our study are extensively used in the field of software defect prediction~\cite{zimmermann2008predicting,zimmermann2009cross,jiarpakdee2019impact,tantithamthavorn2016empirical,rajbahadur2017impact,premraj2011network} and is representative of several corpora and domains. Therefore we argue that our results will still hold. However, future replication across different datasets using our developed methodology might be fruitful. 

Secondly, we only consider one defect prediction context in our study (i.e., within-project defect prediction). Yet, there are multiple defect prediction contexts such as Just-In-Time defect prediction~\cite{hoang2019deepjit,kamei2012large} and cross-project defect prediction~\cite{zimmermann2009cross}. Hence future studies are needed to explore these richer contexts.

Finally, we study a limited number of CS and CA methods and therefore, our results might not readily generalize to other feature importance methods. For instance, there are recent developments like LIME~\cite{ribeiro2016should} that have been proposed in the machine learning community for generating feature importance ranks. Nevertheless, the approach and the metrics that we use in our study are applicable to any feature importance method. Therefore, we invite future studies to use our approach to re-examine our findings on other (current and future) feature importance methods.




%% file: 09-conclusion.tex
\section{Conclusion}
\label{sec:conclusion}

Classifiers are increasingly used to derive insights from data. Typically, insights are generated from the feature importance ranks that are computed by either CS or CA methods. However, the choice between the CS and CA methods to derive those insights remains arbitrary, even for the same classifier. In addition, the choice of the exact feature important method is seldom justified. In other words, several prior studies use feature importance methods interchangeably without any specific rationale, even though different methods compute the feature importance ranks differently. Therefore, in this study, we set out to estimate the extent to which feature importance ranks that are computed by CS and CA methods differ. 

By means of a case study on 18 defect datasets and 6 defect classifiers, we observe that while the computed feature importance ranks by different CA methods strongly agree with each other, the computed feature importance ranks of CS methods do not. Furthermore, the computed feature importance ranks of studied CA and CS methods do not strongly agree with each other either. Except when using different studied CA methods, even the feature reported as the most important feature differs for many of the studied classifiers -- raising concerns about the stability of conclusions across replicated studies. 

We further find that the commonly used defect datasets are rife with feature interactions and these feature interactions impact the computed feature importance ranks of the CS methods (not the CA methods). We also demonstrate that removing these feature interactions, even with simple methods like CFS improves agreement between the computed feature importance ranks of CA and CS methods. We end our study by providing several guidance for future studies, e.g., future research is needed to investigate the impact of advanced feature interaction removal methods on computed feature importance ranks of different CS methods. 

~

\noindent \textbf{ACKNOWLEDGEMENT: }
This research was partially supported by JSPS KAKENHI Japan (Grant Numbers: JP18H03222) and JSPS International Joint Research Program with SNSF (Project ``SENSOR'').

%% file: bios.tex
\vspace{-2cm}
\begin{IEEEbiography}[{\includegraphics[width=1in,height=1.25in,clip,keepaspectratio]{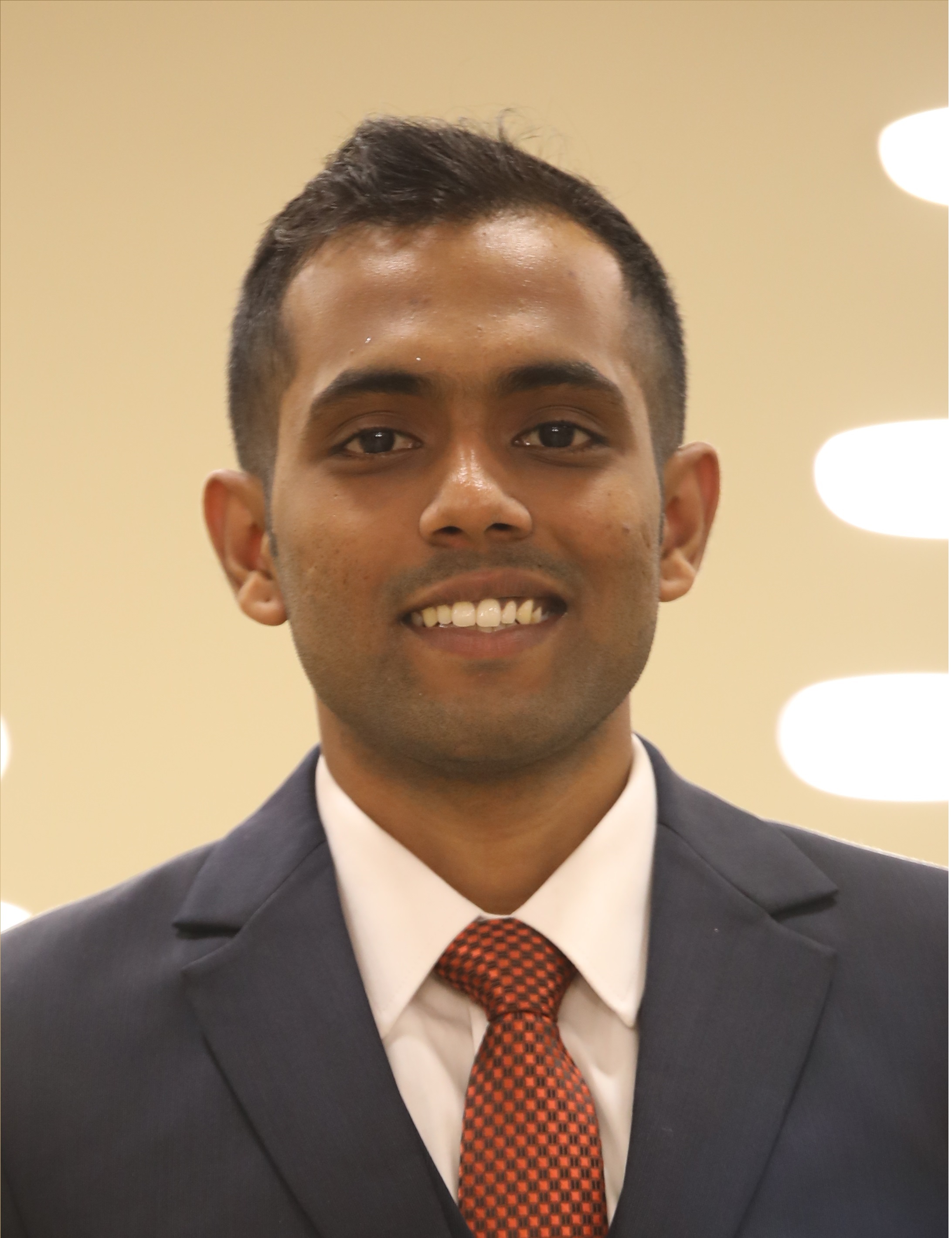}}]{Gopi Krishnan Rajbahadur}
is a Senior Researcher at the Centre for Software Excellence at Huawei, Canada. He holds a PhD in computer science Queen's University, Canada. He received his BE in computer Science and Engineering from SKR Engineering college, Anna University, India. His research interests include Software Engineering for Artificial Intelligence (SE4AI), Artificial Intelligence for Software Engineering (AI4SE), Mining Software Repositories (MSR) and Explainable AI (XAI).
\end{IEEEbiography}

\vspace{-2cm}
\begin{IEEEbiography}[{\includegraphics[width=1in,height=1.25in,clip,keepaspectratio]{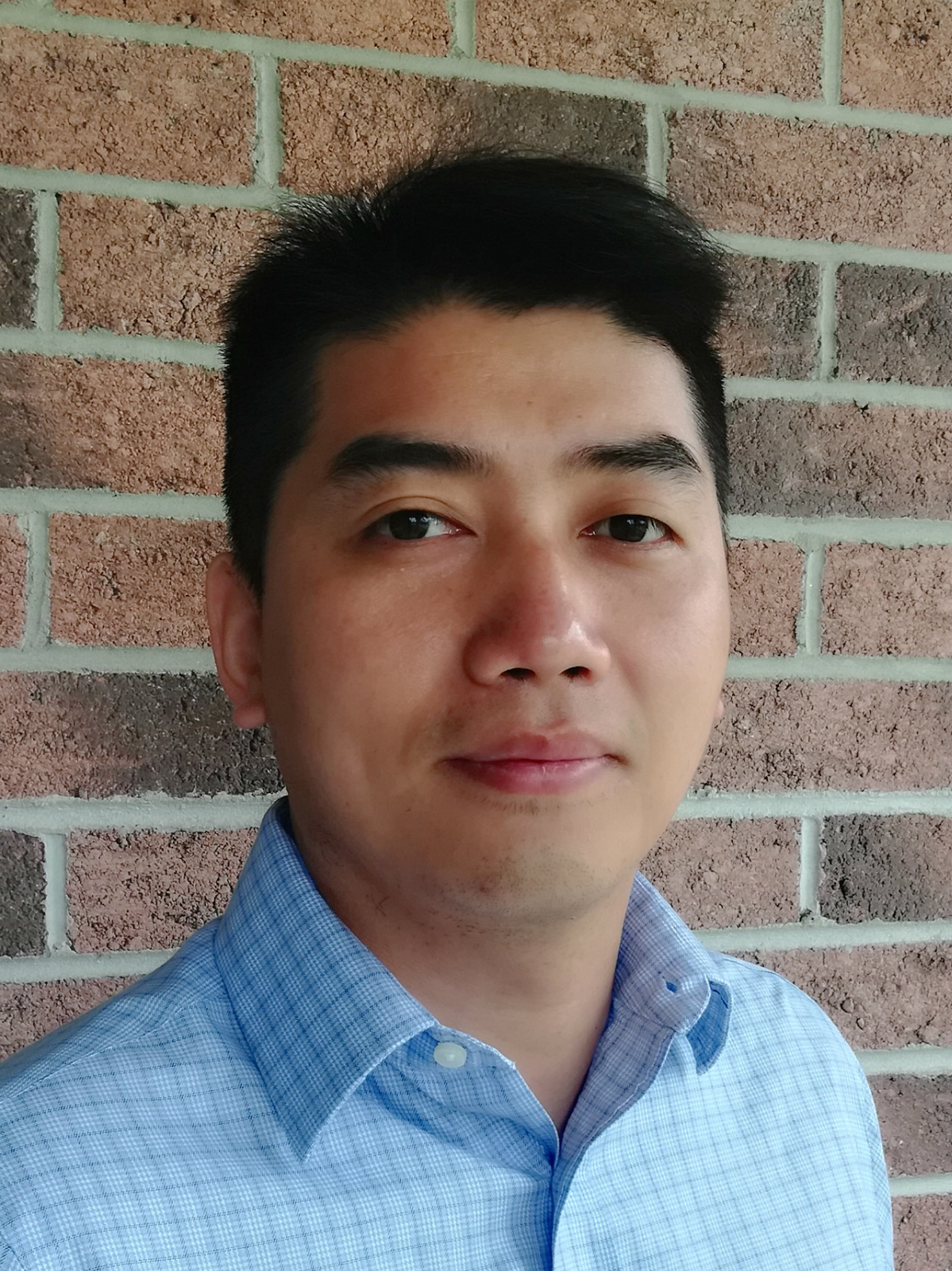}}]{Shaowei Wang}
Shaowei Wang is an assistant professor in the Department of Computer Science at University of Manitoba. He obtained his Ph.D. from Singapore Management University and his BSc from Zhejiang University. His research interests include software engineering, machine learning, data analytics for software engineering, automated debugging, and secure software development. He is one of four recipients of the 2018 distinguished reviewer award for the Springer EMSE (SE's highest impact journal). More information at: \url{https://sites.google.com/site/wswshaoweiwang}.
\end{IEEEbiography}

\vspace{-2cm}

\begin{IEEEbiography}
	[{\includegraphics[width=1in,height=1.25in,clip,keepaspectratio]{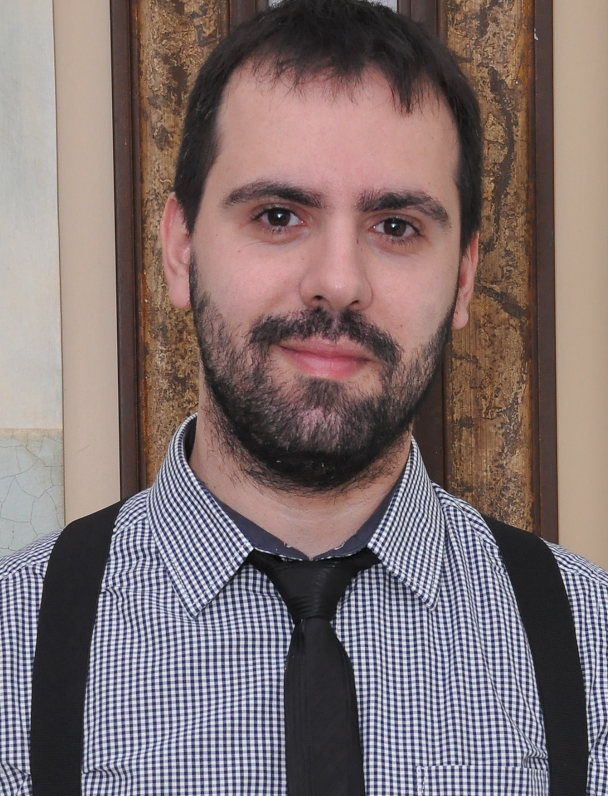}}]
	{Gustavo A. Oliva} is a Research Fellow at the School of Computing of Queen's University in Canada under the supervision of professor Dr. Ahmed E. Hassan. Gustavo leads the blockchain research team at the Software Analysis and Intelligence Lab (SAIL), where he investigates blockchain technologies through the lens of Software Engineering. In addition, Gustavo also conducts research on software ecosystems, explainable AI, and software maintenance. Gustavo received his MSc and PhD degrees from the University of São Paulo (USP) in Brazil under the supervision of professor Dr. Marco Aur{é}lio Gerosa. More information at: \url{https://www.gaoliva.com}
\end{IEEEbiography}

\vspace{-2cm}
\begin{IEEEbiography}[{\includegraphics[width=1in,height=1.25in,clip,keepaspectratio]{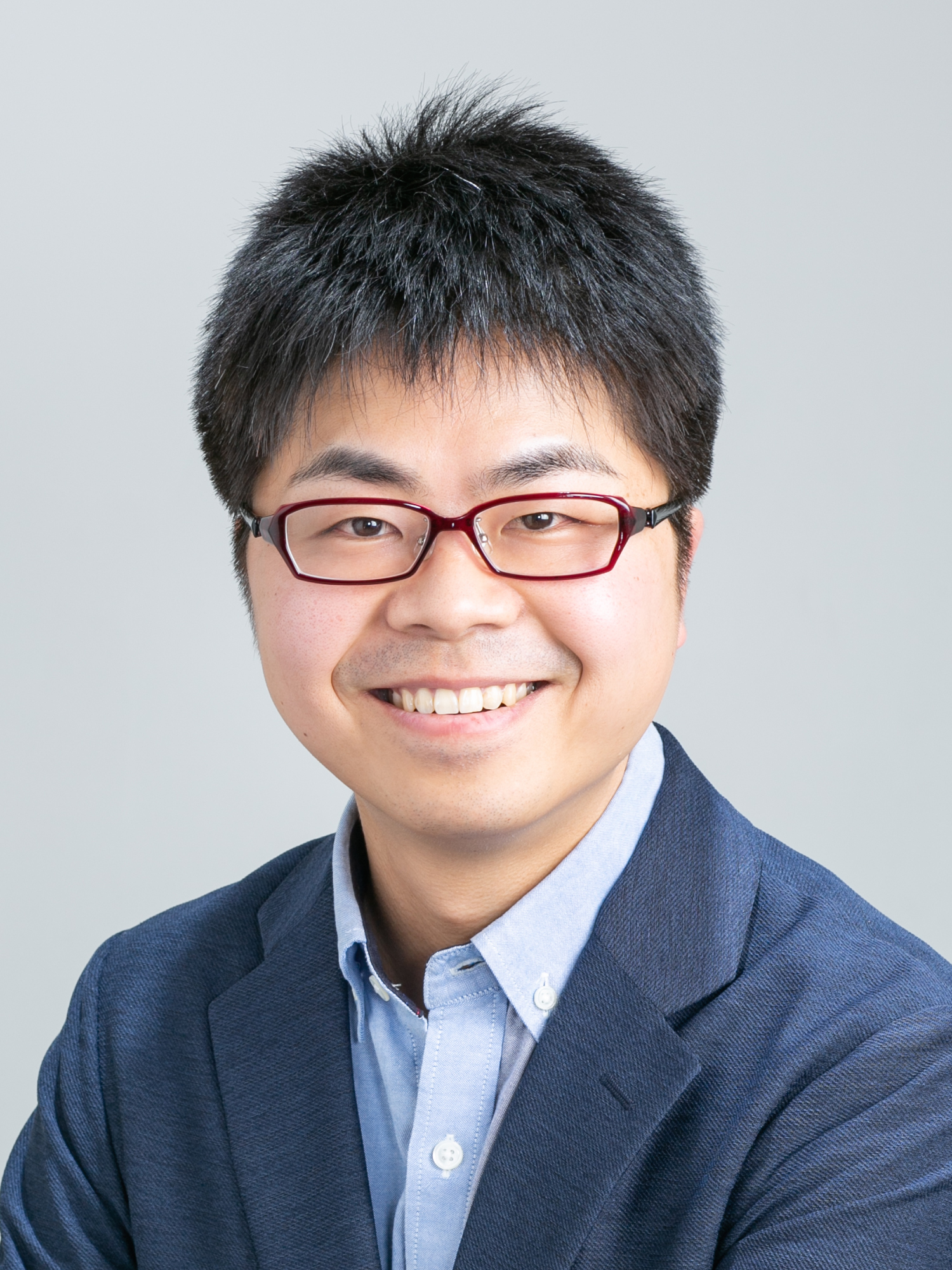}}]{Yasutaka Kamei} is an associate professor at Kyushu University in Japan. He received his BE degree in informatics from Kansai University, and PhD degree in information science from the Nara Institute of Science and Technology. He was a research fellow of the JSPS (PD) from July 2009 to March 2010. From April 2010 to March 2011, he was a postdoctoral fellow at Queen’s University in Canada. His research interests include empirical software engineering, open source software engineering, and mining software repositories (MSR). He serves on the editorial boards of Springer Journal of Empirical Software Engineering. He is a senior member of the IEEE. More information at \url{http://posl.ait.kyushu-u.ac.jp/~kamei}
\end{IEEEbiography}

\vspace{-2cm}
\begin{IEEEbiography}[{\includegraphics[width=1in,height=1.25in,clip,keepaspectratio]{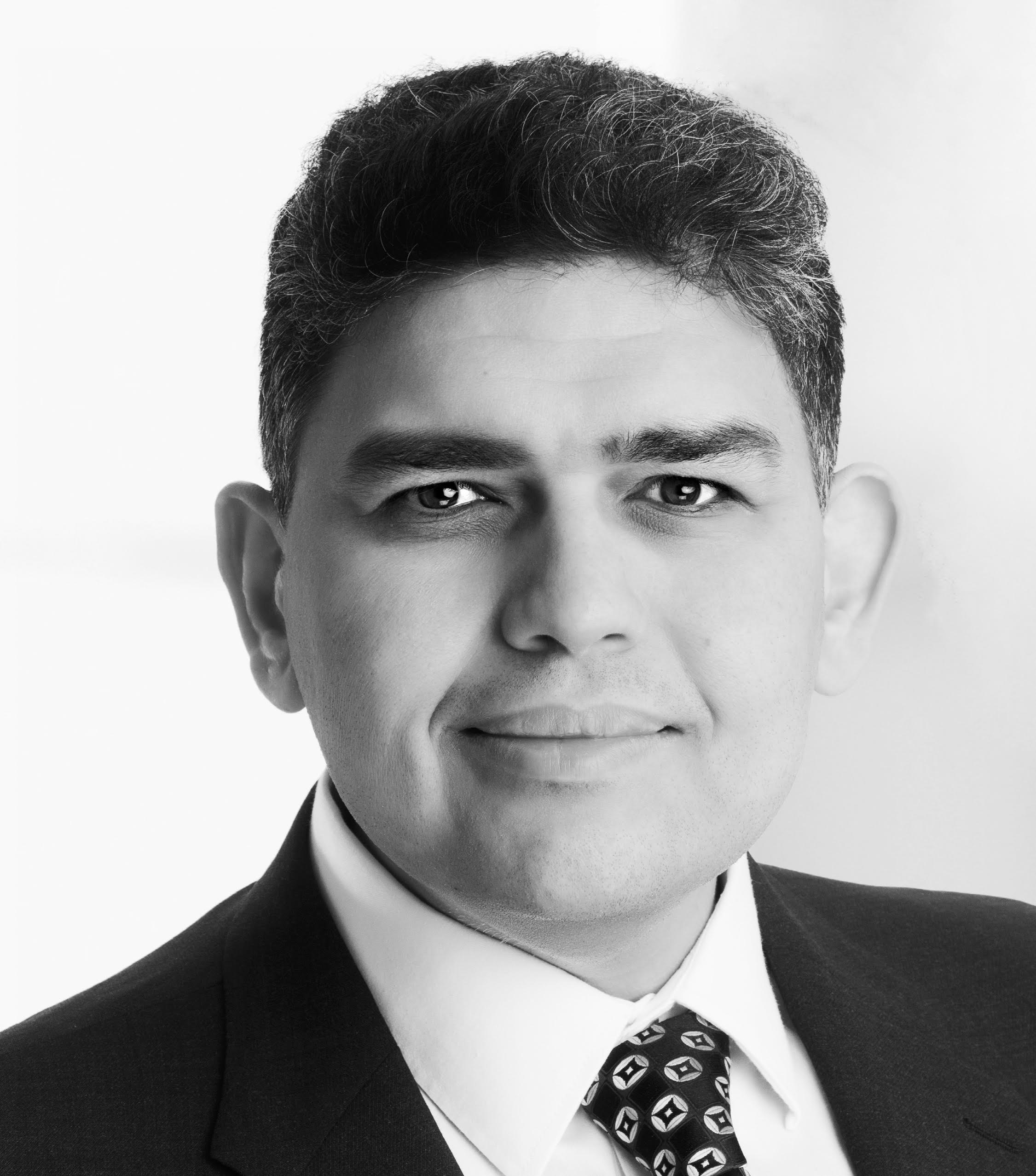}}]{Ahmed E. Hassan} is an IEEE Fellow, an ACM SIGSOFT Influential Educator, an NSERC Steacie Fellow, the Canada Research Chair (CRC) in
Software Analytics, and the NSERC/BlackBerry Software Engineering Chair at the School of Computing at Queen’s University, Canada. His
research interests include mining software repositories, empirical software engineering, load testing, and log mining. He received a PhD in Computer Science from the University of Waterloo. He spearheaded the creation of the Mining Software Repositories (MSR) conference and its research community. He also serves/d on the editorial boards of IEEE Transactions on Software Engineering, Springer Journal of Empirical Software Engineering, and PeerJ Computer Science. More information at: \url{http://sail.cs.queensu.ca}

\end{IEEEbiography}

\clearpage

%% file: appendix.tex
\section{Case Study: Additional Information}

\subsection{Studied Datasets}
\label{appendix:studied-datasets}

Table \ref{tab:data} shows various basic characteristics about each of the studied datasets.

\begin{table}[H]
  \centering
  \caption{Overview of the datasets studied in our case study}
   \begin{threeparttable}
    \begin{tabular}{l|rrrrr}
    \hline
    \textbf{Project} & \multicolumn{1}{l}{\textbf{DR}} & \multicolumn{1}{l}{\textbf{\#Files}} & \multicolumn{1}{l}{\textbf{\#Fts}} & \multicolumn{1}{l}{\textbf{\#FACRA}} & \multicolumn{1}{l}{\textbf{EPV}} \\
    \hline
    Poi-3.0 & 63.5 & 442 & 20 & 12  & 14.05 \\
    Camel-1.2 & 35.53 & 608 & 20 & 10 & 10.8 \\
    Xalan-2.5 & 48.19 & 803 & 20 & 11 & 19.35 \\
    Xalan-2.6 & 46.44 & 885 & 20 & 11 & 20.55 \\
    Eclipse34\_debug & 24.69 & 1065 & 17 & 10 & 15.47 \\
    Eclipse34\_swt & 43.97 & 1485 & 17 & 9  & 38.41 \\
    Pde & 13.96 & 1497 & 15 & 6  & 13.93 \\
    PC5 & 27.53 & 1711 & 38 & 13 & 12.39 \\
    Mylyn & 13.16 & 1862 & 15 & 7  & 16.33 \\
    Eclipse-2.0 & 14.49 & 6729 & 32 & 9  & 30.47 \\
    JM1 & 21.49 & 7782 & 21 & 7  & 79.62 \\
    Eclipse-2.1 & 10.83 & 7888 & 32 & 9  & 26.69 \\
    Prop-5 & 15.25 & 8516 & 20 & 12 & 64.95 \\
    Prop-4 & 9.64 & 8718 & 20 & 12 & 42 \\
    Prop-3 & 11.49 & 10274 & 20 & 12 & 59 \\
    Eclipse-3.0 & 14.8 & 10593 & 32 & 9  & 49 \\
    Prop-1 & 14.82 & 18471 & 20 & 13 & 136.9 \\
    Prop-2 & 10.56 & 23014 & 20 & 13 & 121.55 \\
    \hline
    \end{tabular}%
    \begin{tablenotes}
    \item DR: Defective Ratio, FACRA: Features After Correlation and Redundancy Analysis, Fts: Features
    \end{tablenotes}
    \end{threeparttable}
  \label{tab:data}%
\end{table}%

\subsection{Studied Classifiers}
\label{appendix:classifiers}

Table \ref{tab:approach_classifiers} shows our studied classifiers, the machine learning families to which they belong, and the \textit{caret} function that was used to build the classifiers.

\begin{table}[H]
  \centering
  \caption{Studied classifiers and their hyperparameters}
    \begin{tabular}{p{1.5cm}|p{2cm}|p{1cm}|p{2cm}}
    \hline
    \textbf{Family} & \textbf{Classifier} & \textbf{Caret method} & \textbf{Hyperparameters} \\
    \hline
    Statistical Techniques & Regularized Logistic Regression & glmnet & alpha, lambda \\
    \hline
    Rule-Based Techniques & C5.0 Rule-Based Tree & C5.0Rules & None \\
    \hline
    Neural Networks & Neural Networks (with model averaging) & avNNet & size, decay, bag \\
    \hline
    Decision Trees & Recursive Partitioning and Regression Trees & rpart & K, L, cp \\
    \hline
    Ensemble methods- Bagging & Random Forest & rf & mtry \\
    \hline
    Ensemble methods- Boosting & Extreme Gradient Boosting Trees & xgbTree & nrounds, max\textunderscore depth, eta, gamma, colsample\textunderscore bytree, min\textunderscore child\textunderscore weight, subsample \\
    \hline
    \end{tabular}%
  \label{tab:approach_classifiers}%
\end{table}%

\subsection{Classifier Specific Feature Importance (CS) methods}
\label{appendix:cs-methods}

Table \ref{tab:CS} provides a brief explanation about the inner working of the CS methods that we study in this paper. For a more detailed explanation we refer the readers to~\citet{kuhn2012variable}.

\begin{table*}[!htbp]
  \centering
  \caption{Brief explanation about the working of caret's CS methods that are used in our study.}
    \begin{tabular}{|p{4cm}|p{12cm}|}
    \hline
    \textbf{CS method} & \textbf{Brief explanation} \\
    \hline
    \textbf{Logistic Regression FI (LRFI)} & Classifier coefficient's t-statistic is reported as the feature importance score \\
    \hline
    \textbf{C5.0 Rule-Based Tree FI (CRFI)} & The number of training data points that are covered by the leaf nodes, created from the split of a feature is given as the feature importance score for that feature. For instance, the feature that is split in the root node will have a 100\% importance as all the training samples will be covered by the terminal nodes leading from it. \\
    \hline
    \textbf{Neural Networks (with model averaging) FI (NNFI)} & The feature importance score is given by combining the absolute weights used in the neural network \\
    \hline
    \textbf{Recursive Partitioning and Regression Trees FI (RFI)} & The feature importance score is given by the sum of the reduction in loss function that is brought about by each feature at each split in the tree. \\
    \hline
    \textbf{Random Forest FI (RFFI)} & Average of difference between the Out-of-Bag (OOB) error for each tree in the forest where none of the features are permuted and the OOB error where each of the features is permuted one by one. The feature permutation's impact on the overall OOB error is reported as the feature importance score \\
    \hline
    \textbf{Extreme Gradient Boosting Trees FI (XGFI)} & Feature importance score is given by counting the number of times a feature is used in all the boosting trees of the xgboost tree. \\
    \hline
    \end{tabular}%
    \label{tab:CS}%
\end{table*}%

\subsection{Classifier Agnostic Feature Importance (CA) methods}
\label{appendix:ca-methods}

In the following, we describe the two CA methods that we employ in this study.

\smallskip \noindent\textbf{Permutation feature importance (Permutation).} We use the same permutation feature importance method used by \citet{rajbahadur2017impact}. First, the performance of a classifier is calculated on the non-permuted dataset (we use AUC for measuring the performance in our study). Then each feature in the training set is randomly permuted one by one and this permuted feature along with the other non-permuted features are used to measure the performance of the classifier. If the permutation of a feature decreases the performance of the classifier, compared to the classifier built on the non-permuted dataset, then that feature is considered important. The magnitude of performance changes decides the importance scores of features in a given dataset.

\smallskip  \noindent\textbf{SHapley Additive ExPlanations (SHAP).} We use the method outlined by~\citet{lundberg2017unified}. SHAP uses the game-theory based Shapley values~\cite{shapley1953value}to fairly distribute the credit for a classifier’s output among its features. To do so, for each data point in the train set, SHAP first transforms all the features into a space of simplified binary features. Then SHAP estimates how the output probability of the classifier for the given data point can be expressed as a linear combination of these simplified binary features. These computed feature importance scores are theoretically guaranteed to be optimal, additive, and locally accurate for each data point. Therefore, the overall feature importance scores for each studied feature in the classifier can be given by simply summing the feature importance score of each feature across all the data points in the train set~\cite{covert2020understanding}. For more details about how SHAP computes the feature importance ranks we refer the readers to studies by~\citet{lundberg2017unified,esteves2020understanding}. We use the \textit{vip\footnote{https://cran.r-project.org/web/packages/vip/index.html}} R package for computing the SHAP feature importance scores in our study.

\subsection{Out-of-sample Bootstrap}
\label{appendix:oosb}

The out-of-sample bootstrap is a model validation technique that aims to create a test set that has a similar distribution to that of the training set in each validation iteration. In the following, we briefly describe how it works:

\begin{enumerate}[leftmargin=*]
    \item For a given dataset, N (where N=size of the given dataset) data points are randomly sampled with replacement. These N data points are used as the \textit{train} set for the classifiers.

    \item As the data points were sampled with replacement in the prior step, on an average, 36.8\% of the data points do not appear as a part of the \textit{train} set~\cite{efron1983estimating}. We use these data points as the \textit{test} set for measuring the performance of the constructed classifiers.
\end{enumerate}

The aforementioned process should be repeated $k$ times for each of the studied dataset. Typical values for k include 100, 500, and 1,000.

\subsection{Performance Computation}
\label{appendix:perf-measures}

In the following, we briefly describe each of our adopted performance measures and the rationale for choosing them: 

\smallskip\noindent\textbf{AUC.} The ROC curve plots the True Positive Rate (TPR = TP / (TP + FN)) against the False Positive Rate (FPR = FP / (FP + TN)  for all possible classification thresholds (i.e., from 0 to 1). The use of the AUC measure has several advantages~\cite{ghotra2015revisiting,lessmann2008benchmarking,rajbahadur2019impact,tantithamthavorn2018experience}.  First, AUC does not require one to preemptively select a classification threshold. Hence, the AUC measure prevents our study from being influenced by the choice of classification thresholds~\cite{rajbahadur2017impact}. For instance, when measuring the performance of a classifier with other performance measures like precision and recall, a threshold (e.g., 0.5) needs to be set up to discretize the classification probability into the “Defective” and “Non-Defective” outcome classes. However, setting this threshold is challenging in practice. The AUC measure circumvents this problem by measuring the TPR and FPR at all possible thresholds. Following this, the performance of the classifier is given as the area under this curve. Another commonly used measure to evaluate the performance of a classifier is the Matthews Correlation Coefficient (MCC). We use AUC instead of MCC because (i) MCC is threshold-dependent (similarly to precision and recall) and (ii) there is empirical evidence showing that AUC is generally more discriminative than MCC (even though AUC and MCC tend to be statistically consistent with each other)~\cite{halimu2019empirical}. AUC is robust to imbalanced class~\cite{lessmann2008benchmarking} (even more so than other performance measures like MCC~\cite{halimu2019empirical}). As a consequence, class imbalances that are typically inherent to defect datasets ~\cite{arisholm2006predicting,bird2009fair} are automatically accounted for. The AUC measure ranges from 0 to 1. An AUC measure close to 1 indicates the classifier’s performance is very high. Conversely, an AUC measure close to 0.5 indicates that the classifier’s performance is no better than a random guess.

\smallskip\noindent\textbf{IFA.} The IFA measures the number of false alarms that are encountered before the first defective module is detected by a classifier ~\cite{jiarpakdee2020empirical,huang2017supervised}. We choose the IFA measure in lieu of others (e.g., False Alarm Rate) because several prior studies in fault localization argued that practitioners tend to distrust a classifier whose top recommendations are false alarms. In particular,~\citet{parnin2011automated}, found that developers tend to avoid automated tools if the first few recommendations given by them are false alarms. Therefore, we argue that the number of false alarms that are encountered before the first defective module is detected is a better measure to evaluate the usability of our constructed defect classifiers in practice. The IFA measure is calculated by first sorting the modules in the descending order of their risk as given by the classifiers (i.e, the probability of a module being defective). Then the number of non-defective modules that are predicted as defective before identifying the first true positive (i.e., defective module) is the IFA of a classifier. The IFA measure ranges from 1 (best) to the number of modules in the dataset.

\section{Friedman H-Statistic per dataset}
\label{appendix:friedman}

\begin{table}[H]
  \centering
  \caption{No. of features per dataset with Friedman H-Statistic $\geq$ 0.3 and $\geq$ 0.5 }
  \begin{threeparttable}
    \begin{tabular}{l|r|r}
    \hline
    \textbf{Dataset} & \multicolumn{1}{l|}{\textbf{\#F with H $\geq$ 0.3}} & \multicolumn{1}{l}{\textbf{\#F with H $\geq$ 0.5}}\\
    \hline
    \textbf{Poi-3.0} & 3  & 0 \\
    \textbf{Camel-1.2} & 5 & 4 \\
    \textbf{Xalan-2.5} & 4 & 3 \\
    \textbf{Xalan-2.6} & 2  & 0 \\
    \textbf{Eclipse34\textunderscore debug} & 6  & 3 \\
    \textbf{Eclipse34\textunderscore swt} & 3  & 0 \\
    \textbf{Pde} & 5  & 4 \\
    \textbf{PC5} & 4  & 0 \\
    \textbf{Mylyn} & 4  & 4 \\
    \textbf{Eclipse-2.0} & 6  & 4 \\
    \textbf{JM1} & 7  & 5 \\
    \textbf{Eclipse-2.1} & 6  & 3 \\
    \textbf{Prop-5} & 5  & 4 \\
    \textbf{Prop-4} & 5  & 3 \\
    \textbf{Prop-3} & 5  & 3 \\
    \textbf{Eclipse-3.0} & 6  & 5 \\
    \textbf{Prop-1} & 6  & 2 \\
    \textbf{Prop-2} & 8  & 4 \\
    \hline
    \end{tabular}%
\begin{tablenotes}
    \scriptsize
    \item F - Features
\end{tablenotes}
\end{threeparttable}
  \label{tab:disc2_tab1}%
\end{table}%